\pdfoutput=1

\documentclass[10pt,twocolumn,twoside,journal]{IEEEtran}

\usepackage{cs}
\usepackage{mathsymb}
\usepackage{verbatim}
\usepackage{subfig}
\usepackage{url}
\usepackage{graphicx}
\graphicspath{{./}{../Fig/}{../Figs/}{./Figs/}{./Fig/}}

\usepackage{cite}

\usepackage{algorithm2e}
\let\savedalgorithm\algorithm
\let\savedendalgorithm\endalgorithm
\newenvironment{algorithmic}{%
\savedalgorithm
}{%
\savedendalgorithm
}

\usepackage{algorithm}

\def\paragraph{\textbf}

\usepackage{amsmath}
\usepackage{url}
\usepackage{multirow}
\usepackage{adjustbox}

\def\Integer{\mathbb{Z}^+}

\def\Integer{\mathbb{Z}^+}

\def\brho{{\boldsymbol \rho}}
\def\bvarrho{{\boldsymbol \varrho}}

\def\iprime{{\tau}}

\def\Lag{{\varLambda}}

\def\yiprime{{y_{\iprime}}}

\def\air{{a_{(\iprime|r)}}}
\def\bair{{\ba_{(\iprime|r)}}}
\def\bxi{{\bx_i}}

\def\bxiprime{{\bx_\iprime}}
\def\bxiprimep{{\bx_{\iprime}^{+}}}
\def\bxiprimem{{\bx_{\iprime|r}^{-}}}
\def\myh{{{\tilde{h}}}}

\def\hfraki{ { {\myh}_\iprime} }
\def\hfrakip{ { {\myh}^{+}_\iprime} }
\def\hfrakim{ { {\myh}^{-}_\iprime} }

\def\bpij{{\bp_{(i,j)}}}

\def\bpijp{{\bp_{(i,j)}^{+}}}
\def\bpijm{{\bp_{(i,j)}^{-}}}
\def\bpprime{{\bp^{\prime}}}
\def\rhoir{\rho_{\iprime,r}}
\def\uir{u_{\iprime,r}}
\def\uil{u_{\iprime,l}}

\def\NNr{{\rm NN}_{r}\xspace}

\def\NNyi{{\rm NN}_{y_i}\xspace}
\def\NNnyi{{\rm NN}_{\neq y_i}\xspace}

\def\algonamelong{{\sc C}ompact {\sc B}inary {\sc I}mage {\sc D}escriptor\xspace}
\def\algoname{{\rm C-BID}\xspace}

\def\picodes{{\rm PiCoDes}\xspace}

\def\ClassOne{{Image-based I2C}\xspace}
\def\ClassTwo{{Patch-based I2C}\xspace}

\def\classTwo{{patch-based I2C}\xspace}
\def\ClassOneAb{{{\rm I2C}$^{\rm image}$}\xspace}
\def\ClassTwoAb{{{\rm I2C}$^{\rm patch}$}\xspace}

\def\bbeta{{\boldsymbol \beta}}

\begin{document}

\title{Large-margin Learning of  Compact Binary Image Encodings}

\author{
         Sakrapee Paisitkriangkrai,
         Chunhua Shen,
         Anton van den Hengel
\thanks
{
}
\thanks
{
The authors are with The Australian Center for Visual Technologies,
The University of Adelaide, SA 5005, Australia
(e-mail: \{paul.pais, chunhua.shen, anton.vandenhengel\}@adelaide.edu.au).
Correspondence should be addressed to C. Shen.
}
\thanks
 {
 }
}

\markboth{Manuscript}
{Paisitkriangkrai
\MakeLowercase{\textit{et al.}}: Large-margin Learning of Compact Binary Image Encodings}

\maketitle

\begin{abstract}
The use of high-dimensional features has become a normal practice in
many computer vision applications.
The large dimension of these features is a limiting factor upon the number of data points which may be effectively stored and processed, however.
We address this problem by developing
a novel approach to learning a compact binary encoding,
which exploits both pair-wise proximity and class-label information on training data set.
Exploiting this extra information allows the development of encodings which, although compact,
outperform the original
high-dimensional features in terms of final classification or retrieval performance.
The method is general, in that it is applicable to both
non-parametric and parametric
learning methods.
This generality means that the embedded features are suitable for
a wide variety of computer vision tasks, such as image
classification and content-based image retrieval.
Experimental results demonstrate that the new compact descriptor achieves an
accuracy comparable to, and in some cases better than,
the visual descriptor in the original space despite being significantly more compact.
Moreover, any convex loss function and
convex regularization penalty (\eg, $ \ell_p $ norm with $ p \ge 1 $) can
be incorporated into the framework,
which provides future flexibility.
\end{abstract}


\section{Introduction}
The increasing availability of large volumes of imagery, and the benefits
that have flown from the analysis of large image databases, have seen
significant research effort applied in this area.
The ability to exploit these large data sets, and the range of technologies which may be applied,
is limited by the need to store and process the resulting sets of feature descriptors.
This limitation is particularly visible in retrieval algorithms which must process a
large set of high-dimensional descriptors on-line in response to individual queries.

Effort has been devoted to address these issues and
a hashing based approach has since become the most popular approach
\cite{Andoni2008Near,Liu2012Spline, Weiss2008Spectral}.
It constructs a set of hash functions
that map high-dimensional data samples to low-dimensional binary codes.
The pair-wise Hamming distance between these binary codes can be efficiently computed by using
bit operations.
The new binary descriptor addresses both the issues of efficient data storage
and fast similarity search.
A number of effective hashing methods have been
developed which construct a variety of hash functions.
We divide these works into three categories:
data-independent, data-dependent and object-category-based approaches.
Our paper falls into the last category.
In contrast to all
previous learning algorithms, our approach encodes visual
features of the original high-dimensional space in the form
of compact binary codes, while preserving the underlying
the category-based proximity comparison between the data point in the original space.
To achieve this, we learn hash functions based on a pair-wise distance information
which is presented in a form of triplets.
The proposed approach assigns
a large distance to pairs of irrelevant instances and
a small distance to pairs of relevant instances.
In this paper, we define irrelevant instances to be
samples from different classes and
relevant instances to be samples from the same class.
We formulate our learning
problem in the large-margin framework.
However the number of possible hash functions is infinitely large.
Column generation is thus employed to efficiently solve
this large optimization problem.

The main contributions of this work are as follows.
(i) We propose a novel method (referred to as \algoname)
by which we learn a set of binary output functions (binary
visual descriptors) in a single optimization framework
through the use of the column generation technique.
To our knowledge, our approach is the first learning-based binary descriptor
which exploits both pair-wise similarity and multi-class label information.
By utilizing both neighbourhood and class label information,
the learned descriptor is not only compact but also highly discriminative.
Additionally, our approach is complimentary to many existing computer
vision approaches in significantly reducing the dimension of the data,
that needs to be stored and processed.
(ii) Similar to other column-generation-based algorithms, \eg,
LPBoost \cite{Demiriz2002LPBoost} and PSDBoost \cite{Shen2012Positive},
our approach is robust and highly effective.
Experimental results demonstrate that \algoname not only reduces
the feature storage requirements but also retains or improves upon the classification accuracy
of the original feature descriptors.

\section{Related works}
The method we propose transforms the original
high-dimensional data into a more compact yet highly discriminative feature space.
It is thus a suitable pre-processing step for any computer
vision algorithms, where a massive number of data points are stored
and processed.
Before we propose our approach,
we provide a brief overview and related works on image representation
and image classification.

Learning compact codes or image signatures to represent the
image has been the subject of much recent work.
Compact codes can be categorized into three groups:
data-independent,
data-dependent (unsupervised learning) and
object-category-based approaches.
In the data independent category, compact codes are generated independently of the data.
One of the best known data-independent algorithms of this category is
Locality Sensitive Hashing (LSH) \cite{Datar2004Locality}.
LSH constructs a set of hash functions that maps similar high-dimensional
data to the same low-dimensional binary codes (buckets) with high probability.
These binary codes can then be used to efficiently index the data.
LSH has been used in a wide range of applications such as near-duplicate detection \cite{Ke2004Efficient},
image and audio similarity search \cite{Chum2007Scalable}, and
object recognition \cite{Frome2005Object}.
Since, LSH is data independent, multiple hash tables are often used to ensure it high retrieval accuracy.

Recently, researchers have proposed effective
ways to build data-dependent binary codes in an unsupervised manner.
Examples include Spectral Hashing \cite{Weiss2008Spectral},
Anchor Graph Hashing \cite{Liu2011Hashing},
Spherical Hashing \cite{Heo2012Spherical}
and Spline Regression Hashing \cite{Liu2012Spline}.
These algorithms learn compact binary codes
which aim to preserve the
pair-wise distances between input data.
The authors then either solve the exact optimization problem
or an approximate solution to the original problem.
The learned hash functions enforce the hamming distance
between codewords to approximate the actual distance between the data
in the original space.
Other data-dependent descriptors also include bag of visual words (BOW),
which is a sparse vector of occurrence counts of local features
\cite{Lazebnik2006Beyond} and vector of locally aggregated
descriptors (VLAD), which aggregates local descriptors
into a vector of fixed dimension.

Finally, object-category-based approaches aim to learn a compact binary descriptor
in a supervised manner.
Torresani \etal represent a compact image code as a bit-vector which
are the outputs of a large set of classifiers \cite{Torresani2010Efficient}.
Li \etal propose a high-level image representation
as the response map from a large number of pre-trained generic
visual detectors \cite{Li2010Bank}.
Bergamo \etal propose a \picodes descriptor which learns a
binary descriptor by directly minimizing a multi-class classification
objective \cite{Bergamo2011PICODES}.
The descriptor has been shown to outperform many hashing algorithms and achieves
state-of-the-art results at various descriptor sizes.
One of the drawbacks of this approach, however, is that it can take several weeks to learn an encoding.
Later, the same authors proposed the  more efficient
meta-class (MC) descriptor \cite{Bergamo2012MetaClass},
which adopts the label tree learning algorithm of \cite{Bengio2010Label}.
The final descriptor is a concatenation of all classifiers learned using label trees.
The descriptor size is fixed.
In contrast to all previous learning algorithms,
the proposed approach encodes visual features of the
original high-dimensional space in the form of compact image signatures
based on Image-to-Class distance \cite{Boiman2008Defense}.
The resulted feature is not only more compact but also
preserves the underlying proximity comparison between the
data point in the original space.

Image classification can be categorized into a parametric and
non-parametric method.
A parametric method constructs image features as a vector of predefined
length.
A discriminative classifier, such as SVM, is used to learn the decision function
that best separates the training data of the $k$-th class from other training
samples.
The most well known example is a bag of visual words model (BoW).
Visual features are first extracted at multiple scales.
These raw features are quantized into a set of visual words.
A new representative feature is then calculated on the basis of a histogram of the visual words.
This technique often reduces the high dimensional feature space to just
a few thousand visual words.
The BoW approach underpins  state-of-the-art results in
many image classification problems\cite{
Boureau2010Learning, Lazebnik2006Beyond, Yang2009Linear, Wang2010Locality}.

In the second approach, the non-parametric method,
local features belonging to the same class
are grouped together to represent that specific class.
Image-to-Class (I2C) distance from a given test image to each class is defined
as the sum of distance between every local feature
in the test image and its nearest neighbour in each class.
This approach is also known as a patch-based Naive Bayes
Nearest Neighbour model (NBNN) \cite{Boiman2008Defense}.
In contrast to the BoW approach, the NBNN based approach does not
quantize visual descriptors but it relies on nearest neighbour search
over image patches.
The classification is performed based on the summation of Euclidean
distances between local patches of the query image and
local patches from reference classes
(hence the name Image-to-Class distance).
I2C distance has demonstrated promising results
on several image data sets when experimented with linear
distance coding \cite{Wang2013Linear}.
Several researchers have attempted to improve the performance of NBNN.
Behmo \etal corrected NBNN for the case in which there are unbalanced training sets \cite{Behmo2010Towards}.
Tuytelaars \etal proposed a kernel NBNN which uses NBNN response
features as the input features from which to learn the second layer using a kernel
SVM \cite{Tuytelaars2011NBNN}.
McCann and Lowe proposed to speed up
the effectiveness and efficiency of NBNN by merging patches from all
classes into a single search structure \cite{McCann2012Local}.
Wang \etal proposed a per-class Mahalanobis distance metric to enhance
the performance of I2C distance for small number of local features \cite{Wang2010Image}.

Note that the practicality of both BoW and NBNN approaches are limited
by the fact that they require highly distinctive feature descriptors
for reasonable results.
Unfortunately, feature distinctiveness often
comes at the cost of increased database size.
In this work, we propose a novel \algonamelong (\algoname) which
addresses this shortcoming.
The proposed approach is applicable to both parametric and
non-parametric image classification frameworks.

\begin{figure*}[t]
    \begin{center}
        \subfloat[\ClassTwoAb distance] {
          \includegraphics[width=0.48\textwidth,clip]{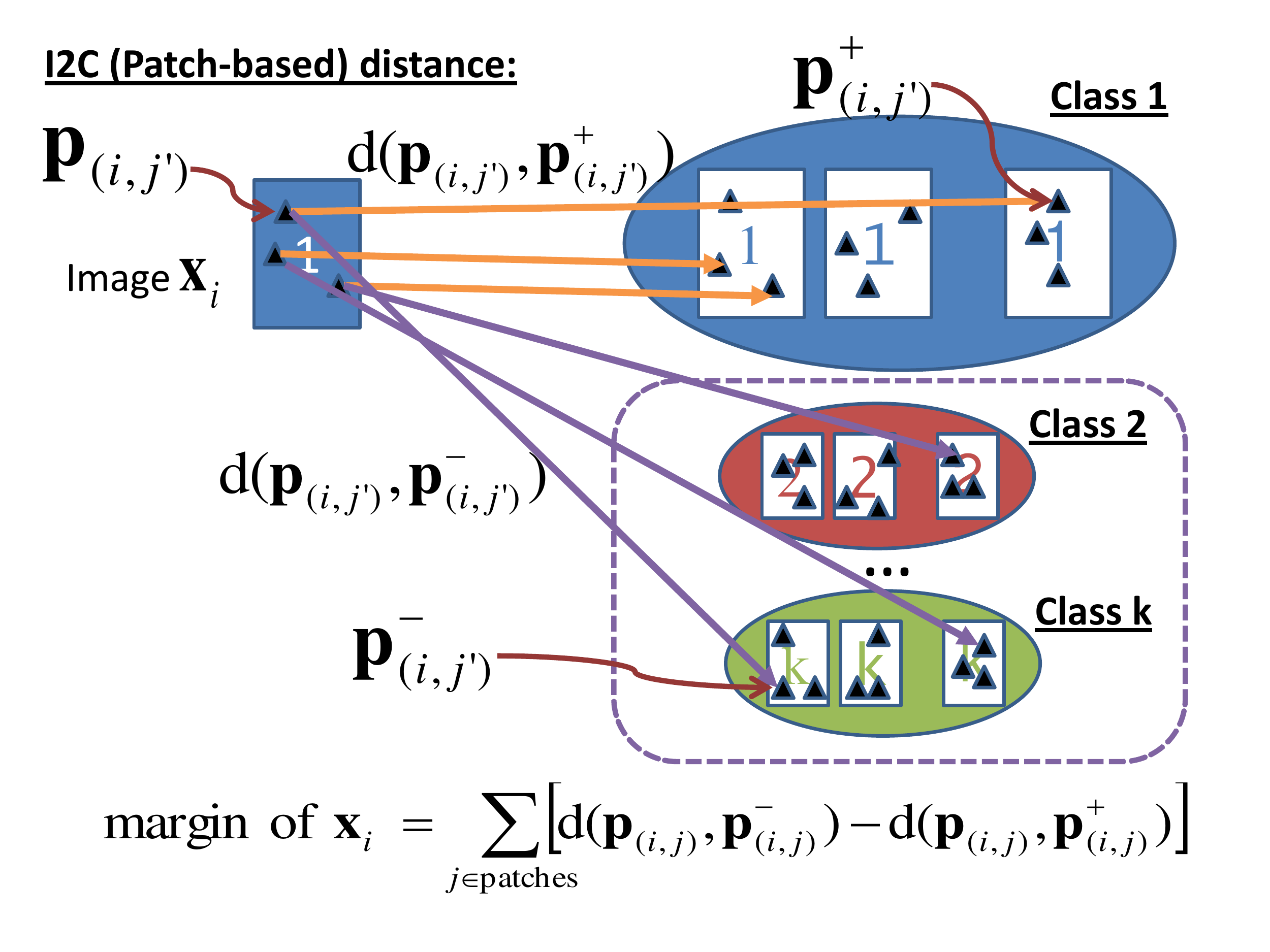}
          \label{fig:illus_patch}
        }
        \subfloat[\ClassOneAb distance] {
          \includegraphics[width=0.48\textwidth,clip]{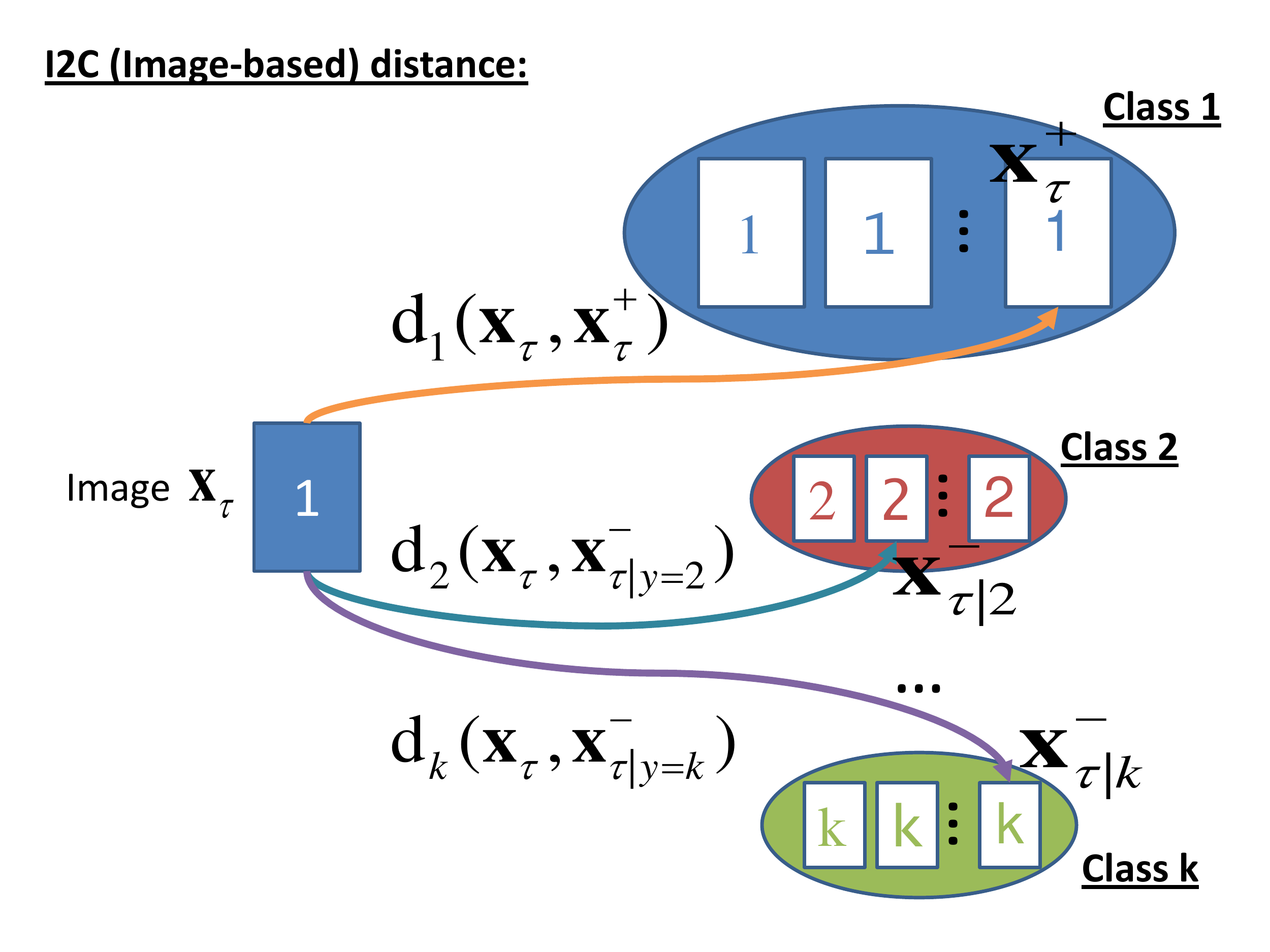}
          \label{fig:illus_image}
        }
    \end{center}
    \caption{
    Illustration of \ClassTwoAb and \ClassOneAb distances.
    Here we assume that the image $\bx_i$ belongs to class $1$.
    \textbf{(a)}
    Patches in the image are represented by black triangles.
    Each orange arrow maps the patch $\bp_{(i,j^{\prime})}$ to the closest patch
    with the same class label $\bp_{(i,j^{\prime})}^{+}$.
    Each purple arrow maps the patch $\bp_{(i,j^{\prime})}$ to the closest patch
    with the different class label $\bp_{(i,j^{\prime})}^{-}$.
    To achieve high accuracy, we prefer $d(\bpij,\bpijp)$ to be small and
    $d(\bpij, \bpijm)$ to be large.
    Our method learns hash functions that preserve the relative comparison
    relationships in the data, \ie,
    the margin of $\bx_i$ should be as large as possible.
    \textbf{(b)}
    Each image now contains one single patch.
    Each orange arrow maps the image $\bx_i$ to
    several closest images with the same class label $\bx_i^{+}$.
    }
    \label{fig:illustration}
\end{figure*}

\section{Approach}

In this section, we present the I2C distance definition.
We then define our margins and propose two approaches which learn a set of hash functions
and their coefficients (weighted Hamming distance).
We then discuss the application and computational complexity of the proposed approach.

\subsection{Background}

\paragraph{I2C distance}
Suppose we are given a set of training images
$\{ (\bx_i, y_i) \}_{i=1}^m$
where $\bxi \in \mathbb{R}^{d}$ represents a $d$-dimensional image
and $y_i \in \{1,\cdots,k\}$ the corresponding class label.
Here $m$ is the number of training instances and
$k$ is the number of classes.
Let $\{ \bp_{(i,1)}, \cdots, \bp_{(i,n)} \}$
denote a collection of local feature descriptors,
in which $\bpij$ represents features extracted from the
$j$-th patch in the $i$-th image.
Here $n$ represents the number of patches in $\bxi$ and
visual features can simply be its pixel intensity values
or local distinct feature descriptors such as SIFT \cite{Lowe2004Distinctive},
Edge-SIFT \cite{Zhang2013Edge} or SURF \cite{Bay2006SURF}.
To calculate the I2C distance from an image $\bxi$ to a candidate
class $r$, NBNN finds the nearest neighbour to each feature $\bpij$
from each class $r$.
The I2C distance is defined as the
sum of Euclidean distances between each feature $\bpij$ in image $\bxi$
and its nearest neighbour from class $r$, $\NNr(\bpij)$.
The distance can be written as \cite{Boiman2008Defense}:
$\sum_{j=1}^n \| \bpij - \NNr(\bpij) \|^2$.
The NBNN classifier is of the form
\begin{align}
F(\bxi) = \argmin_{r=1,\cdots,k} \sum_{j=1}^n \| \bpij - \NNr(\bpij) \|^2
\end{align}
where $\NNr(\bpij)$ is the nearest neighbour patch of $\bpij$ in class $r$.

\paragraph{Patch-based I2C margin}
The objective of this paper is to learn a set of
compact and highly discriminative binary codes $\{h_1(\cdot),\ldots,h_t(\cdot)\}$,
$\in \cal{H}$, each of which maps the patch $\bpij$
into the binary space
$\{-1,+1\}$.
The I2C margin is based on
the I2C distance proposed for the NBNN classifier.
To better model a pair-wise proximity, we define the weighted distance between
any two input patches, $\bp$ and $\bpprime$ as
$d(\bp, \bpprime) =  \sum_{s=1}^t w_{s} | h_s (\bp) - h_s (\bpprime) |$.
Here $\bw = \left[ w_{1}, \cdots, w_{t} \right] \in \mathbb{R}^{t}$  is the
(non-negative) %
weight vector.
We formulate the I2C margin of the $i$-th image based on the intuition
that the distance of patch $\bpij$
to any other classes ($\bpijm$)
should be larger than the distance
to its belonging class ($\bpijp$).
The \ClassTwoAb margin can be written as\footnote{The margin definition defined here has also been used in various literatures, \eg, feature selection \cite{Kira1992Practical, Sun2010Local}, classification and metric learning \cite{Weinberger2006Distance, Shen2012Positive},
data embedding based on similarity triplets \cite{Maaten2012Stochastic}, \etc
},
\begin{align}
\varrho_{i} &= \sum_{j=1}^n \left[ d(\bpij, \bpijm) - d(\bpij, \bpijp) \right] \\ \notag
 &= \left( \textstyle \sum_j \ba_{(i,j)} \right) \bw
\end{align}
where $\bpijm = \NNnyi(\bpij)$ is the nearest neighbour patch of $\bpij$ with a
different class label (purple arrows in Fig.~\ref{fig:illus_patch}),
$\bpijp = \NNyi(\bpij)$ is the nearest neighbour patch of $\bpij$
with the same class label (orange arrows in Fig.~\ref{fig:illus_patch}),
$\ba_{(i,j)} = \left[ a_{(i,j),1}, \cdots, a_{(i,j),t} \right]$, and
$a_{(i,j),s} =  \bigl[ | h_s (\bpij) - h_s (\bpijm) | - | h_s (\bpij) - h_s (\bpijp) | \bigr]$.
We thus aim to learn a set of functions, $\{h_s (\cdot) \}_{s=1}^{t}$,
where $h(\bp; \bbeta, b) = \sign( \bbeta^\T \bp + b )$, such that
the weighted distance between two binary codes
$[ h_1 (\bpij), \cdots, h_t (\bpij)]$ and
$[ h_1 (\bpijp), \cdots, h_t (\bpijp)]$ remains small.

\paragraph{Image-based I2C margin}
In this paper, we extend the I2C margin originally designed for key-point matching
to image matching.
We represent the whole image as a large single patch.
Similar to the previous margin, the orange arrow
can be mapped to multiple patches (images).
We illustrate our new \ClassOneAb in Fig.~\ref{fig:illus_image}.
We define the new margin
based on the intuition that the distance between
a pair of irrelevant images
(image $\bxiprime$ and images from any
other classes) should be larger than
the distance between a pair of relevant images
(image $\bxiprime$ and images from its belonging class).
The training data can be written in a form
of triplets $\{ (\bxiprime, \bxiprimep, \bxiprimem) \}_{\iprime=1}^{| \mathcal{S} |}$
where each triplet indicates the relationship of three images,
$\mathcal{S}$ is the triplet index set
and $| \mathcal{S} |$ is the total number of triplets.
Here $\bxiprimep$ is a set of nearest neighbours from the same class
and $\bxiprimem$ is a set of nearest neighbours from class $r$ ($r \neq \yiprime$).
Unlike the \ClassTwoAb distance, we parameterized each class by a weight vector
associated with the class label.
Since we have $k$ classes, we define the matrix $\bW = [\bw_1, \cdots, \bw_k]$,
such that the $r$-th column of $\bW$ contains Hamming weights for
class $r$.
The weighted Hamming distance between any two data points, $\bxi$
and $\bx_j$, is defined as
$\Delta_r(\bxi, \bx_j) =  \sum_{s=1}^t w_{s,r} | h_s (\bxi) - h_s (\bx_j) |$.
Here $\left[ w_{1,r}, \cdots, w_{t,r} \right] \in \mathbb{R}^{t}$  is the
(non-negative) %
weight vector associated with the class label $r$.
We define the margin for $\bxiprime$ as the difference between
two distances (i) weighted distance between $\bxiprime$ and $\bxiprimem$,
and (ii) weighted distance between $\bxiprime$ and $\bxiprimep$:
\begin{align}
  \label{EQ:margin1}
  \rho_{\iprime,r} &= \Delta(\bxiprime, \bxiprimem) - \Delta(\bxiprime, \bxiprimep) \\  \notag
           &= \bair \bw_{\yiprime} - \bair \bw_{r}
\end{align}
where $\iprime$ is an index of the triplet set
$\{ (\bxiprime, \bxiprimep, \bxiprimem) \}_{\iprime=1}^\mathcal{|S|}$,
$\bair = \left[ a_{(\iprime|r),1}, \cdots, a_{(\iprime|r),t} \right]$, and
$a_{(\iprime|r),s} = \bigl[ | h_s (\bxiprime) - h_s (\bxiprimem) | - | h_s (\bxiprime) - h_s (\bxiprimep) | \bigr]$.
Similar to the patch-based approach, we thus aim to learn a
set of functions, $\{h_s (\cdot) \}_{s=1}^{t}$,
where $h(\bx; \bbeta, b) = \sign( \bbeta^\T \bx + b )$.

\subsection{Learning compact descriptors}
\label{sec:approach}

In this section, we design the new approach using the large margin framework.
The optimization problem is formulated such that the distance of
the nearest instance from the same class (\emph{nearest hit})
is smaller than the distance of the nearest instance from other classes
(\emph{nearest miss}).

\paragraph{\ClassTwo optimization}
The general $\ell_1$-regularized optimization problem for the \classTwo margin is
\begin{align}
\label{EQ:patch_a}
    \min_{ \bw, \bvarrho }   \;
    &
    {\textstyle \sum}_{i=1}^{m} L (\varrho_{i}) + \nu  \| \bw \|_1     \\ \notag
    \st \; &
    \varrho_{i} = \left( {\textstyle \sum}_j \ba_{(i,j)} \right) \bw,
    \forall i,j; \; \bw \geq 0, \notag
\end{align}
where $L$ can be any convex loss function,
the regularization parameter $\nu$ determines
the trade-off between the data-fitting loss function
and the model complexity,
subscripts $i$ and $j$ index images and patches, respectively.
Here we introduce the auxiliary variables, $\bvarrho$, to obtain a
meaningful dual formulation.
For the logistic loss, the learning problem can be expressed as:
    \begin{align}
    \label{EQ:patch_b}
        \min_{ \bw, \bvarrho }   \;
        &
        {\textstyle \sum}_{i} \log \left( 1 + \exp(-\varrho_{i}) \right)
          + \nu  \| \bw \|_1     \\ \notag
        \st \; &
        \varrho_{i} = \left( {\textstyle \sum}_j \ba_{(i,j)} \right) \bw,
    \forall i,j; \; \bw \geq 0, \notag
    \end{align}
The Lagrangian of \eqref{EQ:patch_b} can be written as:
\begin{align}
    \notag
    \Lag \; & = \sum_{i}
        \log ( 1 + \exp( - \varrho_{i}) ) +
        \nu \b1^\T \bw \\ \notag
      & \quad
      -\sum_{i} v_i   \left(
      \varrho_{i} - \left( {\textstyle \sum}_j \ba_{(i,j)} \right) \bw
      \right) - \bq^\T \bw,
     \notag
\end{align}
with $\bq \geq 0$.
The dual function is:
\begin{align}
    \notag
    \inf_{\bw,\bvarrho} \Lag \; & =
        \inf_{\bw,\bvarrho} {\textstyle \sum}_{i} \log ( 1 + \exp( - \varrho_{i}) )
        - {\textstyle \sum}_{i} v_i \varrho_i \\ \notag
        & \quad \quad +  \overbrace{ \bigl( {\textstyle \sum}_{i} v_i
                 ( {\textstyle \sum}_j \ba_{(i,j)} )
                - \bq^\T + \nu \b1^\T \bigr)   }^\text{must be zero} \bw.     \notag
\end{align}
Since the convex conjugate function of the logistic loss,
$\log \left( 1 + \exp( -x ) \right)$, is
$(-u) \log (-u) + (1+u) \log (1+u)$ if $-1 \leq u \leq 0$
and $\infty$ otherwise.
The Lagrange dual for the logistic loss is:
\begin{align}
    \notag
        \max_{ \bv }  %
        &
        - {\textstyle  \sum}_{i=1}^{m}
             \Bigl[ - v_i \log ( -v_i ) + (1 + v_i)
            \log ( 1 + v_i ) \Bigr]
         \\ \notag
        \st \;
        &
        {\textstyle \sum}_{i=1}^{m}  v_i
        \left(  {\textstyle \sum}_j \ba_{(i,j)} \right) \geq - \nu \b1^\T, \;
        \bv \geq 0.
\end{align}
By reversing the sing of $\bv$, we obtain:
\begin{align}
    \label{EQ:dual_margin2}
        \min_{ \bv }  %
        &
        {\textstyle  \sum}_{i=1}^{m}
            \Bigl[ v_i \log ( v_i ) + (1 - v_i)
            \log ( 1 - v_i ) \Bigr]
         \\ \notag
        \st \;
        &
        {\textstyle \sum}_{i=1}^{m}  v_i
        \left(  {\textstyle \sum}_j \ba_{(i,j)} \right) \leq \nu \b1^\T, \;
        \bv \geq 0.
\end{align}

\paragraph{\ClassOne optimization}
The general $\ell_1$-regularized optimization problem
we want to solve is
    \begin{align}
    \label{EQ:image_a}
        \min_{ \bW, \brho }   \;
        &
        {\textstyle \sum}_{\iprime,r} L (\rho_{\iprime,r}) + \nu  \| \bW \|_1     \\ \notag
        \st \; &
        \rho_{\iprime,r} = \bair \bw_{\yiprime} - \bair \bw_{r},
        \forall \iprime, r; \; \bW \geq 0. \notag
    \end{align}
The learning problem for the logistic loss can be expressed as:
    \begin{align}
    \label{EQ:image_b}
        \min_{ \bW, \brho }   \;
        &
        {\textstyle \sum}_{\iprime,r} \log \left( 1 + \exp(-\rho_{\iprime,r}) \right)
          + \nu  \| \bW \|_1     \\ \notag
        \st \; &
        \rho_{\iprime,r} = \bair \bw_{\yiprime} - \bair \bw_{r},
        \forall \iprime, r; \; \bW \geq 0. \notag
    \end{align}
The Lagrangian of \eqref{EQ:image_b} can be written as
\begin{align}
    \notag
    \Lag(& \bW,\brho, \bU, \bZ) \; = {\textstyle \sum}_{\iprime,r}
        \log ( 1 + \exp( - \rho_{\iprime,r}) ) +
        \nu {\textstyle \sum}_{r} \bw_{r} \\ \notag
      &- {\textstyle \sum}_{\iprime,r} \uir   (
      \rho_{\iprime,r} - \bair \bw_{\yiprime}
      + \bair \bw_{r} )
      - \trace(\bZ^\T \bW),
\end{align}
with $\bZ \geq 0$.
The Lagrangian function is
\begin{align}
     \inf_{\bW,\brho}  \Lag (&\bW,\brho,\bU,\bZ) \\ \notag
  = &- {\textstyle \sum}_{\iprime,r} \sup_{\rhoir} \Bigl( \uir \rhoir - \log(1+\exp(-\rhoir)) \Bigr) \\ \notag
    &\quad + \inf_{\bW} \Bigl(
        \nu {\textstyle \sum}_{r} \bw_{r}
         + {\textstyle \sum}_{\iprime,r} \uir \bair \bw_{\yiprime} \\ \notag
    &\quad \quad \quad    - {\textstyle \sum}_{\iprime,r} \uir \bair \bw_{r}
         - \trace(\bZ^\T \bW)   \Bigr).
\end{align}
At optimum the first derivative of the Lagrangian with respect to
each row of $\bW$ must be zeros, \ie,
$\frac{\partial \Lag}{\partial \bw_r } =  {\bf 0}$, and therefore
\begin{align}
    \notag
    \;
    \sum_{\iprime \mid \yiprime=r}
        &\left( \textstyle \sum_{l}  \uil  \right) \bair
        -  \sum_{\iprime} \uir \bair
        = \bz_{r:} - \nu \b1^\T  \\ \notag
    &\Rightarrow
    {\textstyle \sum}_{\iprime} \delta_{r,\yiprime} \left( \textstyle \sum_{l} \uil \right) \bair
    - {\textstyle \sum}_{\iprime} \uir  \bair
    \geq  -\nu \b1^\T,
\end{align}
$ \forany r $ and $\delta_{s,t} = 1$ if $s = t$ and
$0$, otherwise.
The Lagrange dual problem is
\begin{align}
   \notag
        \max_{ \bU }   \quad
        &
        - {\textstyle \sum}_{\iprime,r}
            \Bigl[ -\uir \log \left( -\uir \right)
            + \left(1 + \uir \right)
            \log\left( 1 + \uir \right) \Bigr]
         \\ \notag
        \st \quad &
        {\textstyle \sum}_{\iprime} \left[ \delta_{r,\yiprime}  (\textstyle \sum_{l} \uil)
          - \uir \right] \bair \geq  -\nu \b1^\T, \forall r;
\end{align}
By reversing the sign of $\bU$, we obtain
\begin{align}
        \notag
        \min_{ \bU }   \quad
        &
        {\textstyle \sum}_{\iprime,r}
            \Bigl[ \uir \log \left( \uir \right)
            + \left(1 - \uir \right)
            \log\left( 1 - \uir \right) \Bigr]
         \\ \label{EQ:image_dual}
        \st \quad &
        {\textstyle \sum}_{\iprime} \left[ \delta_{r,\yiprime}  (\textstyle \sum_{l} \uil )
          - \uir \right] \bair \leq  \nu \b1^\T, \forall r;
\end{align}

\paragraph{General convex loss}
In this section, we generalize our approach to any convex losses
with $\ell_1$-norm penalty.
Note that our approach is not limited to the $\ell_1$-norm regularized framework
but other $\ell_p$-norm penalties ($p > 1$) can also be applied (see the Appendix).
The Lagrangian of \eqref{EQ:image_a} can be written
as\footnote{The Lagrange dual of \eqref{EQ:patch_a} can be formulated similarly.}:
\begin{align}
    \notag
    \Lag = {\textstyle \sum}_{\iprime,r} &L (\rho_{\iprime,r}) +
        \nu {\textstyle \sum}_{r} \bw_{r}
      -{\textstyle \sum}_{\iprime,r} \uir  \\ \notag
      &(\rho_{\iprime,r} - \bair  \bw_{\yiprime}
      + \bair \bw_{r} )
      - \trace(\bZ^\T \bW),
\end{align}
with $\bZ \geq 0$.
Following our derivation for the logistic loss, the Lagrange dual can be written as
\begin{align}
    \label{EQ:general_dual}
        \min_{ \bU, \bQ }   \quad
        &
        {\textstyle \sum}_{\iprime,r} L^{\ast} ( - \uir )     \\ \notag
        \st \quad &
        {\textstyle \sum}_{\iprime} \left[ \delta_{r,\yiprime}  (\textstyle \sum_{l} \uil )
          - \uir \right] \bair \leq  \nu \b1^\T, \forall r;
\end{align}
where $L^{\ast}(\cdot)$ is the Fenchel dual function of
$L(\cdot)$.
Through the KKT optimality condition, the duality gap between the
solutions of the primal optimization problem \eqref{EQ:image_a} and
the dual problem \eqref{EQ:general_dual} must coincide since
both problems are feasible and the Slater's condition is satisfied.
The required
relationship between the optimal values of $\bU$ and $\brho$
(for \ClassOneAb) and between the optimal values of $\bv$ and $\bvarrho$
(for \ClassTwoAb) thus hold at optimality.
These relationships can be expressed as
$\uir = - L^{\prime} (\rhoir)$ (for \ClassOneAb) and
$v_i = - L^{\prime} (\varrho_i)$ (for \ClassTwoAb).
For the logistic loss of \eqref{EQ:image_a} and \eqref{EQ:patch_a},
we can write these relationships as:
\begin{align}
    \label{EQ:KKT_a}
    \uir^\ast = \frac{ \exp(-\rhoir^\ast) }
        {   1+\exp(-\rhoir^\ast) }
\end{align}
and
\begin{align}
    \label{EQ:KKT_b}
    v_i^{\ast} = \frac{ \exp(-\varrho_i^\ast) }
        {   1+\exp(-\varrho_i^\ast) },
\end{align}
respectively.

\paragraph{Learning binary output functions}
Since there may be infinitely many constraints in \eqref{EQ:general_dual},
we use column generation
to identify an optimal set of constraints\footnote{
Note that constraints in the dual correspond to variables in the primal.} \cite{Demiriz2002LPBoost}.
Column generation allows us to avoid solving the original problem, which has a large number of constraints, and instead to consider a much smaller problem which
guarantees the new solution to be optimal for the original problem.
The algorithm begins by finding the most violated dual constraint in the dual problem \eqref{EQ:general_dual} and inserts this constraint into the new optimization problem (which corresponds to inserting a primal variable).
From \eqref{EQ:general_dual} the subproblem for generating the most violated dual constraint, $h^{\ast}(\cdot)$, is
\begin{align}
\label{EQ:weaklearner}
    h^{\ast}(\cdot) = \argmax_{h(\cdot)\in \mathcal{H}, r}
    \;
    {\textstyle \sum}_{\iprime} ( \delta_{r,\yiprime} (\textstyle \sum_{l} \uil ) - \uir ) \air,
\end{align}
where $\air = | h (\bxiprime) - h (\bxiprimem) | - | h (\bxiprime) - h (\bxiprimep) |$
and $h(\bxi) = \sign( \bbeta^\T \bxi + b )$.
At each iteration, we thus add an additional constraint to the dual problem.
The process continues until there are no violated constraints
or the maximum number of iterations is reached.

Not only \eqref{EQ:weaklearner} is non-convex but also exhaustively evaluating
all possible candidates in the hypothesis space is infeasible.
We thus treat the problem of learning binary output functions as the gradient
ascent optimization problem.
To achieve this, we replace $|\cdot|$ and $\sign(\cdot)$ operators in \eqref{EQ:weaklearner}
with $(\cdot)^{2}$ and $\frac{2}{\pi} \arctan(\cdot)$, respectively,
due to their differentiability.
Note that it is possible to replace $\sign(\cdot)$ with other sigmoid functions, \eg, $\frac{1}{1+\exp(-t)}$.
We used $\arctan(\cdot)$ here as it has
been successfully applied to learn features in \cite{Liu2007Gradient}.
To summarize, we replace $a_{(\iprime)}$ and $h(\bxiprime; \bbeta, b)$
in \eqref{EQ:weaklearner}
with,
\begin{align}
    \notag
    a_{(\iprime)}
    \rightarrow
    \bigl( h  (\bxiprime ) - h (\bxiprimem ) \bigr)^2 - &
        \bigl( h  (\bxiprime ) - h  (\bxiprimep )  \bigr)^2, \\ \notag
    h(\bxiprime; \bbeta, b)
    \rightarrow \frac{2}{\pi} & \arctan ( \bbeta^\T \bxiprime + b ).
\end{align}
Given an initial value of $(\bbeta, b)$, we iteratively search for the new value
that leads to the larger value of
$\sum_{\iprime} ( \delta_{r,\yiprime} (\textstyle \sum_{l} \uil ) - \uir ) \air $
based on the gradient ascent method.
To find a local maximum,
we repeatedly take a step proportional to
the gradient of the function
at the current point.
The iteration stops when the magnitude of the gradient is smaller than
some threshold, \ie, the objective value is at its local maximum.

For ease of exposition, we define the following variables:
$\hfraki = h(\bxiprime)$, $\hfrakip = h (\bxiprimep)$,
$\hfrakim = h(\bxiprimem)$ and
$\omega_{(\iprime,r)} = (\delta_{r,\yiprime}  (\textstyle \sum_{l} \uil ) - \uir)$.
Let $\Pi = \sum_{\iprime} ( \delta_{r,\yiprime} (\textstyle \sum_{l} \uil ) - \uir ) \air$.
The gradient of $\Pi$ can then be expressed as:
\begin{align}
    \label{EQ:dldbeta}
    \frac{\partial \Pi }{\partial \bbeta} &= {\textstyle \sum}_{\iprime} \omega_{(\iprime,r)}
            \frac{\partial \air}{\partial \bbeta}, \\
    \label{EQ:dldb}
    \frac{\partial \Pi }{\partial b} &= {\textstyle \sum}_{\iprime} \omega_{(\iprime,r)}
            \frac{\partial \air }{\partial b}, \\ \notag
\end{align}
where
\begin{align}
    \notag
    \frac{\partial \air }{\partial \bbeta} &=
        2 (\hfraki - \hfrakim) \left( \frac{\partial \hfraki}{\partial \bbeta} -
                                \frac{\partial \hfrakim}{\partial \bbeta} \right)
        \\ \notag
        &
        \quad
        - 2 (\hfraki - \hfrakip) \left( \frac{\partial \hfraki}{\partial \bbeta} -
                                \frac{\partial \hfrakip}{\partial \bbeta} \right),
        \\ \notag
    \frac{\partial \hfraki}{\partial \bbeta} &= \frac{2}{\pi}
                    \frac{ \bxiprime }{1 + (\bbeta^\T \bxiprime + b)^2}, \\ \notag
    \frac{\partial \air }{\partial b} &=
        2 (\hfraki - \hfrakim) \left( \frac{\partial \hfraki}{\partial b} -
                                \frac{\partial \hfrakim}{\partial b} \right)
        \\ \notag
        &
        \quad
        - 2 (\hfraki - \hfrakip) \left( \frac{\partial \hfraki}{\partial \bbeta} -
                                \frac{\partial \hfrakip}{\partial \bbeta} \right),
        \\ \notag
    \frac{\partial \hfraki}{\partial b} &= \frac{2}{\pi}
                    \frac{ 1 }{1 + (\bbeta^\T \bxiprime + b)^2}.
\end{align}

Since using a fixed small learning rate can yield a poor convergence,
a more sophisticated off-the-shelf optimization method can be a better alternative.
In our implementation, we use a limited memory BFGS (L-BFGS) \cite{Zhu1997Algorithm}
to perform a line search algorithm.
At each L-BFGS iteration, we use \eqref{EQ:dldbeta} and \eqref{EQ:dldb}
to compute the search direction.
Since \eqref{EQ:weaklearner} is highly non-convex, choosing the initial value of $(\bbeta,b)$ is
critical to finding the global maximum.
We randomly generate a large number of $(\bbeta,b)$ pairs and
use the pair with the highest response as an initial value for L-BFGS.
Note that the same solver has also been used to speed up training in deep learning \cite{Le2011Optimization}.

\SetKwInput{KwInit}{Initialize}

\begin{algorithm}[t]
\caption{Training algorithm for \ClassTwoAb \algoname.
}
\begin{algorithmic}
\small{
   \KwIn{
    \\1)      Training examples $\{\bx_i,y_i\}_{i=1}^m$
              and their associated patches $\{ \bpij \}$;
     $
     \;
     $
     \\ 2)    The maximum number of bits, $t$;
   }

   \KwOut{
      \\ 1) $t$-bit binary functions, $\{h_s (\cdot)\}_{s=1}^t$;
      \\ 2) Associated Hamming weights, $\bw$;
}

\KwInit {
   \\1)      $s \leftarrow 0$;
   $ \; $
   \\2)      Initialize dual variables, $v_i = 1/m$;
}

\While{ $s < t$ }
{
  1) Train a new binary function by solving
  $ h_s(\cdot) = \argmax_{h(\cdot)\in \mathcal{H}}
    \;
    {\textstyle \sum}_{i=1}^{m}  v_i
        \left(  {\textstyle \sum}_j \ba_{(i,j)} \right) $ \;
  2) Add the new binary function, $h_s(\cdot)$, into the current set \;
  3) Solve the primal problem \eqref{EQ:patch_a}  \;
  4) Update dual variables using \eqref{EQ:KKT_b} \;
  5) $s \leftarrow s + 1$ \;}
} %
\end{algorithmic}
\label{ALG:alg_patch}
\end{algorithm}

\paragraph{Learning Hamming weights}
Hamming weights can be calculated in a totally corrective manner as in \cite{Demiriz2002LPBoost}.
For the logistic loss formulation, the primal problem has $tk$ variables
and $\mathcal{|S|}$ constraints.
Here $ |\cdot| $ counts the number of elements in the triplet set.
Since the scale of the primal problem is much smaller than the dual
problem\footnote{
Not only is $\mathcal{|S|} \gg k$
but we also have more training samples than the number
of binary output functions to be learned, \ie, one usually trains
$1024$ binary output functions (to generate $1024$-bit codes)
but the number of training samples could be larger
than $10,000$.},
we solve the primal problem instead of the dual problem.
The primal problem, \eqref{EQ:image_a}, can be solved using an
efficient Quasi-Newton method like L-BFGS-B, and the dual variables
can be obtained using the KKT condition, \eqref{EQ:KKT_a} and \eqref{EQ:KKT_b}.
The details of our \ClassTwoAb and \ClassOneAb \algoname algorithms
are given in Algorithm~\ref{ALG:alg_patch} and \ref{ALG:alg_image}, respectively.

\SetKwInput{KwInit}{Initialize}

\begin{algorithm}[t]
\caption{Training algorithm for \ClassOneAb \algoname.
}
\begin{algorithmic}
\small{
   \KwIn{
    \\1)    A set of training data in the form of triplets
      $\{ (\bxiprime, \bxiprimep, \bxiprimem) \}_{\iprime=1}^{ | \mathcal{S} | }$
     $
     \;
     $
     \\ 2)    The maximum number of bits, $t$;
   }

   \KwOut{
      \\ 1) $t$-bit binary functions, $\{h_s (\cdot)\}_{s=1}^t$;
      \\ 2) Associated Hamming weights for each class, $\{ \bw_{r} \}_{r=1}^k$;
}

\KwInit {
   \\1)      $s \leftarrow 0$;
   $ \; $
   \\2)      Initialize dual variables, $\uir = 1/( | \mathcal{S} | \cdot r)$;
}

\While{ $s < t$ }
{
  1) Train a new binary function,
  $h_s(\cdot) = $
  $ \argmax_{h(\cdot) \in \mathcal{H} ,r}  \Pi (\bbeta,b)$,
  using the gradient ascent technique (see text)\;
  2) Add the new binary function, $h_s(\cdot)$, into the current set \;
  3) Solve the primal problem \eqref{EQ:image_a}  \;
  4) Update dual variables using \eqref{EQ:KKT_a} \;
  5) $s \leftarrow s + 1$ \;}
} %
\end{algorithmic}
\label{ALG:alg_image}
\end{algorithm}

\subsection{Application}
In this section, we illustrate how the proposed compact binary code can be
applied to the image classification problem using both
\ClassTwoAb and \ClassOneAb approaches.

\paragraph{\ClassTwo approach}
In NBNN the training data for a class is a collection of
small patches densely extracted from given training images\cite{Boiman2008Defense}.
Since the dimension of visual descriptors is large and patches are
sampled densely,
the storage space and the retrieval time are two critical issues that
need to be addressed.
In this section, we illustrate how our approach can be applied
to reduce the storage space and improve the retrieval time.
Following NBNN, we first extract a large number of visual patches
from each training image at the same orientation but multiple spatial scales.
During training, each patch is mapped to have the same binary code as
the most similar patch from the same class.
During evaluation, a large number of patches are extracted from the test image and
the label is assigned based on the minimal distance between test patches and training patches, \ie,
$\argmin_{\ell} \, \sum_{j=1}^{n} \sum_{s=1}^{t} w_s
| h_s(\bp_j) - h_s(  \bp_{j+} ) |$,
where $n$ is the total number of patches in the image and
$\bp_{j+}$ represents the training patch from class $\ell$
such that
$d \bigl( [h_1(\bp_j),\cdots,h_t(\bp_j)], [h_1(\bp_{j+}),\cdots,h_t(\bp_{j+})] \bigr)$
$= \sum_{s=1}^{t} w_s
| h_s(\bp_j) - h_s(  \bp_{j+} ) |$ is minimal.
Note that,
in order to further improve the efficiency during retrieval, binary outputs
of the training data, \ie, $\{h_s (\bpij) \}_{s=1}^{t}$, $i = 1, \cdots, m$,
$j = 1, \cdots, n$,
can be pre-computed offline.
The details of our \algoname classifier for \classTwo distance are given in Algorithm~\ref{ALG:alg3}.
\SetKwInput{KwInit}{Initilaize}

\begin{algorithm}[t]
\caption{\ClassTwoAb \algoname classifier.
}
\begin{algorithmic}
\small{
   \KwIn{
      \\ 1) Query image, $I$;
      \\ 2) $t$-bit binary functions, $\{h_s (\cdot)\}_{s=1}^t$;
      \\ 3) Hamming weights, $\{ w_{s} \}_{s=1}^k$;
   }
   \KwOut{
      Predicted label, $\ell^{\ast}$;
    }
  1)
  Compute visual descriptors for each patch, $\{ \bp_{j} \}_{j=1}^n$,
  from the test image \;
  2)
  $\ell^{\ast} = \argmin_{\ell} \,
  \sum_{j=1}^{n}  \sum_{s=1}^{t}
  w_s | h_s(\bp_j) - h_s(  \bp_{j+} ) |$ \\
  where $\bp_{j+}$ is the nearest patch to $\bp_j$ in the new binary space,
  \ie, $d \bigl( [h_1(\bp_j),\cdots,h_t(\bp_j)], [h_1(\bp_{j+}),\cdots,h_t(\bp_{j+})] \bigr)$
  is minimal,
  and $\bp_{j+}$ is from class $\ell$ \; 
} %
\end{algorithmic}
\label{ALG:alg3}
\end{algorithm}

\paragraph{\ClassOne approach}
The typical steps involved in the image-based model are: (i) feature extraction
(ii) learning a multi-class classifier.
Our algorithm falls in between step (i) and (ii).
In other words, we represent the output of step (i) in a more compact form.
The new descriptor significantly reduces the storage space required
while preserving pair-wise distances between the input data.

\subsection{Computational complexity}
During training (offline), we need to iteratively find the
best hash function \eqref{EQ:weaklearner} and solve the primal problem \eqref{EQ:image_a}.
The computational complexity during training is similar to LPBoost \cite{Demiriz2002LPBoost}.
During evaluation (online), the time complexity is almost the same as other
hashing algorithms.
With a proper implementation, weighted hamming distance can be calculated
almost as fast as hamming distance.
For example, one can use the hamming distance to retrieve similar items.
Retrieved items can be re-ranked using hamming weights.
To further improve the speed, the weighted hamming distance between
$8$ consecutive bits of the query and binary codes can be pre-computed
and stored in a lookup table \cite{Norouzi2012Hamming}.
In addition, vector multiplication can be computed efficiently
using a linear algebra library.

\section{Experiments}
\label{sec:exp}

We evaluate the proposed \algoname on both synthetic data set
and computer vision tasks.
Unless otherwise specified, we use the logistic loss with $\ell_1$-norm
penalty.

\subsection{\ClassTwo distance}
In this section, we perform experiments using our \algoname with NBNN based approach.
All images are resized to a maximum of $150 \times 150$ pixels prior to processing.
We use dense SIFT features in all algorithms.
We randomly select $10$ images per class as the training set and
$25$ images per class as the test set.
All experiments are repeated $5$ times and the average accuracy  reported.
For Caltech-101, we use five classes similar to \cite{Behmo2010Towards}: faces, airplanes, cars-side, motorbikes and background.
In this experiment, we use $1$ nearest neighbour from the same class
and one nearest neighbour from each other classes.
We compare our approach with
BoW with an additive kernel Map of \cite{Vedaldi2010Efficient},
the original NBNN with SIFT features \cite{Boiman2008Defense},
NBNN with an approximate NN (ANN) search using best-bin first search of
kd-tree\footnote{We use a C-MEX implementation of \cite{Vedaldi08Vlfeat}
and limit the maximum number of comparisons to $500$.},
NBNN with LSH\footnote{We use a C-MEX implementation of
\cite{Aly2011Indexing} with $25$ hash tables ($20$ hash functions per
table), $\ell_2$-norm distance and the size of the bin for
hash functions is set to $0.25$.},
and NBNN with SIFT (PCA) features\footnote{PCA is performed to reduce
the size of the original SIFT descriptor.
The projected data is stored as a floating point ($32$ bits).}
and report experimental results in Table~\ref{tab:exp_NBNN}.
From the table, our approach consistently outperforms
all NBNN-like approaches
and {\em performs best on average}.
We suspect that the performance improvement of \algoname is due to
the large margin learning and
the introducing of non-linear functions
into the original SIFT descriptor.
From the table, our algorithm at $128$ bits consistently performs better than SIFT (PCA)
at $512$ bits.
From the same table, ANN search tradeoffs speed over accuracy.

\paragraph{Storage}
For BoW, we divide an image into $20$ sub-images
and the number of visual words is $600$.
Hence there are $36,000$ dimensional features
being extracted from each image.
We also map each feature vector to a higher-dimensional feature space via
an explicit map: $\mathbb{R}^n \rightarrow \mathbb{R}^{n(2r+1)}$
where $r$ is set to $1$ \cite{Vedaldi2010Efficient}.
Each descriptor is stored as a floating point ($32$ bits),
each image requires approximately $141$ kB of storage.

For NBNN, we extract dense SIFT at $4$ different scales.
Assuming that a total of $10,000$ SIFT features are extracted and kept in
the database from each image (as a $32$-bit floating point representation),
NBNN requires approximately $5,000$ kB of storage per image.
Using our approach with $32$ binary output functions,
the original $128$ dimensional SIFT descriptor ($512$ bytes)
is now reduced to $4$ bytes
(a $128$-fold reduction in storage) and
each image now takes up approximately $39$ kB of storage.

Our approach is also {\em much more efficient than
the original NBNN} due to its compact binary codes and the use of
bitwise exclusive-or operation.
Since ANN algorithms, \eg, best bin first and LSH, only index the data,
original data points still need
to be kept in order to determine the exact distance between the query patch
and patches kept in the database.
Hence the storage of ANN is similar to that of the exact NN.

\paragraph{Evaluation time}
We also report the average evaluation time of various NBNN based approaches
(excluding feature extraction time and index building time)
in the last row of Table~\ref{tab:exp_NBNN}.
Our approach has a much lower evaluation time than the original NBNN with SIFT
due to more compact binary codes and the use of bitwise exclusive-or
operation.
For LSH, we set the number of hash functions per hash table to $20$.
We use $25$ hash tables.
Hence there are a total of $500$ hash functions ($500$ binary
descriptors).
However, LSH has a much higher evaluation time than our approach.
We suspect that most of the evaluation time is spent in calculating the
distance between training samples that are mapped to the same hash bucket,
\ie, hash collision.

\begin{table*}[tb]
  \caption{Experiments on \classTwo distance.
      Performance comparison between
      BoW + kernel map + SVM \cite{Vedaldi2010Efficient},
      SIFT + NBNN \cite{Boiman2008Defense},
      SIFT + NBNN + kd-tree search, SIFT + NBNN + LSH,
      SIFT (PCA) + NBNN and
      our approach on various data sets.
      The average accuracy and standard deviation (in percentage)
      are reported.
      The last row reports the average evaluation time per image per class
      of NBNN based approaches (excluding feature extraction time).
      Experiments are executed on a single core of Intel Core
      i$7$ $930$ ($2.8$ GHz) with $12$ GB memory.
      {\em Our approach achieves the higest average accuracy
      while having a much lower storage cost
      and evaluation time.}
  }
  \centering
  \scalebox{0.95}
  {
  \begin{tabular}{l|c|c|c|c|cc|ccc}
  \hline
   \multirow{2}{*}{} & BoW + & NBNN &
   NBNN & NBNN &
   \multicolumn{2}{c|}{NBNN + SIFT (PCA)} & \multicolumn{3}{c}{\algoname} \\
  \cline{6-10}
  & Kernel Map & Exact NN & Best bin first & LSH & $256$ & $512$ bits &
  $16$ & $32$ & $128$ bits \\
  \hline
  \hline
  Graz-02 & $70.67$ ($5.7$) & $70.40$ ($7.9$) & $67.47$ ($6.0$) &  $62.93$ ($5.3$) &
    $66.13$ ($6.6$) & $68.80$ ($8.3$) &  %
    $65.87$ ($11.4$) & $67.73$ ($9.4$) & $\mathbf{72.27}$ ($\mathbf{7.4}$) \\
  Caltech-101 & $\mathbf{86.88}$ ($\mathbf{2.2}$) & $84.96$ ($4.0$) &
    $84.64$ ($3.9$) & $84.80$ ($3.0$) &
    $74.72$ ($3.6$) & $83.04$ ($4.9$) &
    $73.12$ ($7.0$) & $80.96$ ($4.8$) & $86.24$ ($4.2$) \\
  Sports-8 & $57.40$ ($3.1$) & $62.20$ ($3.4$) &
    $59.60$ ($4.1$) & $51.70$ ($3.6$)&
    $48.40$ ($5.2$) & $54.70$ ($3.9$) &
    $42.70$ ($3.4$) & $56.90$ ($3.4$) & $\mathbf{63.00}$ ($\mathbf{4.3}$) \\
  Scene-15 &  $55.73$ ($2.4$) & $54.72$ ($2.6$) &
    $53.97$ ($2.1$) & $40.21$ ($4.7$) &
    $46.24$ ($2.1$) & $54.67$ ($1.8$) &
    $30.03$ ($1.9$) & $54.51$ ($1.8$) & $\mathbf{57.28}$ ($\mathbf{2.8}$) \\
  \hline
  Avg. accuracy & $67.67$ & $68.07$ & $66.42$ & $59.91$ &
    $58.87$ & $65.30$ &
    $52.93$ & $65.03$ & $\mathbf{69.70}$ \\
  Storage$/$image & $141.0$ kB
    & $5,000.0$ kB & $5,000.0$ kB & $5,000.0$ kB &
    $312.5$ kB & $625.0$ kB &
    $19.5$ kB & $39.1$ kB & $156.3$ kB \\
  Eval.$/$img$/$class &
     & $78.8$ sec. & $1.55$ sec. & $3.83$ sec.&
    $5.50$ sec. & $8.62$ sec. &
    $0.15$ sec. & $0.24$ sec. & $1.11$ sec.\\
  \hline
  \end{tabular}
  }
  \label{tab:exp_NBNN}
\end{table*}

\subsection{\ClassOne distance}

\begin{figure*}[t]
    \begin{center}
        \includegraphics[width=0.24\textwidth,clip]{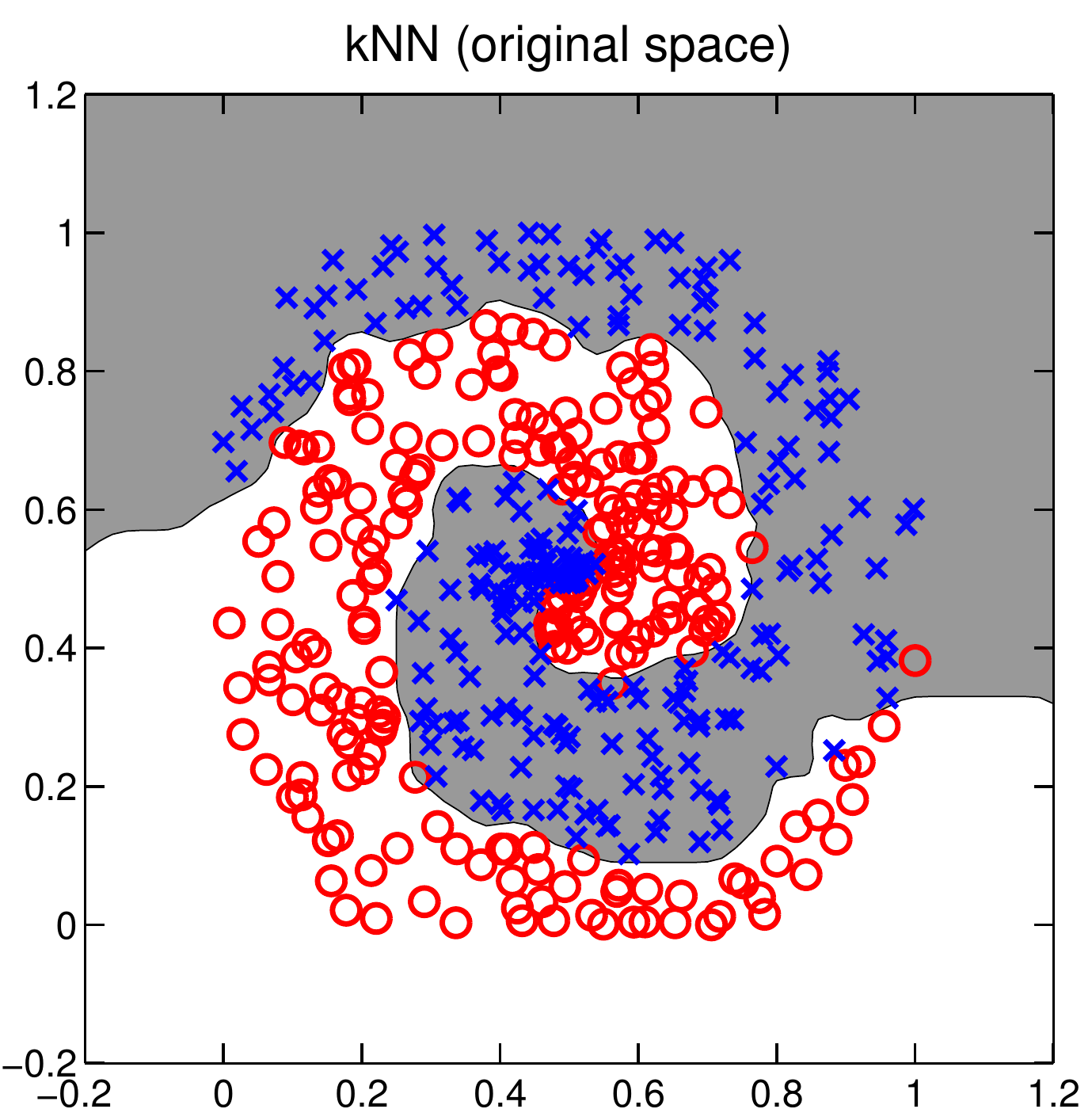}
        \includegraphics[width=0.24\textwidth,clip]{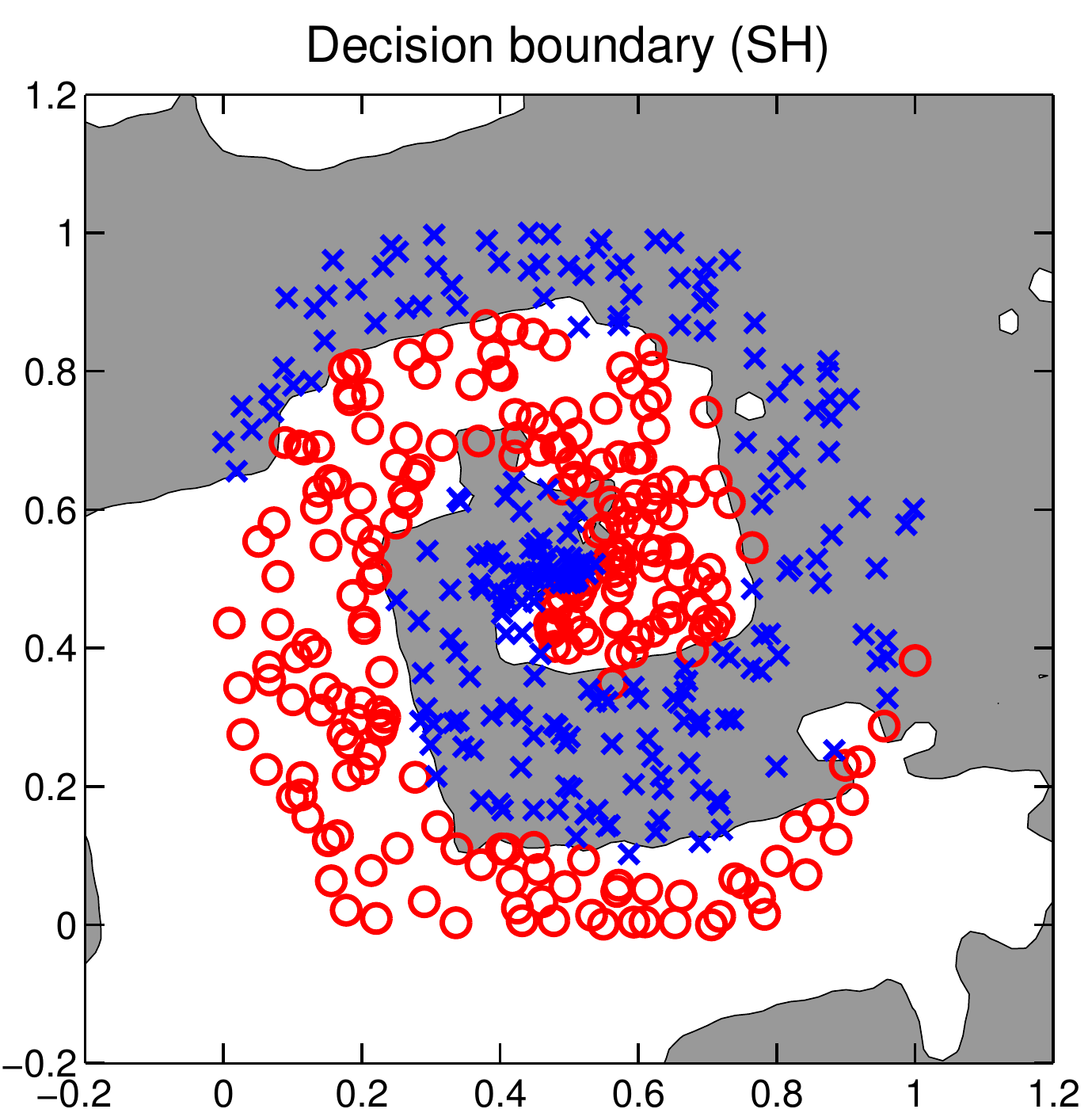}
        \includegraphics[width=0.24\textwidth,clip]{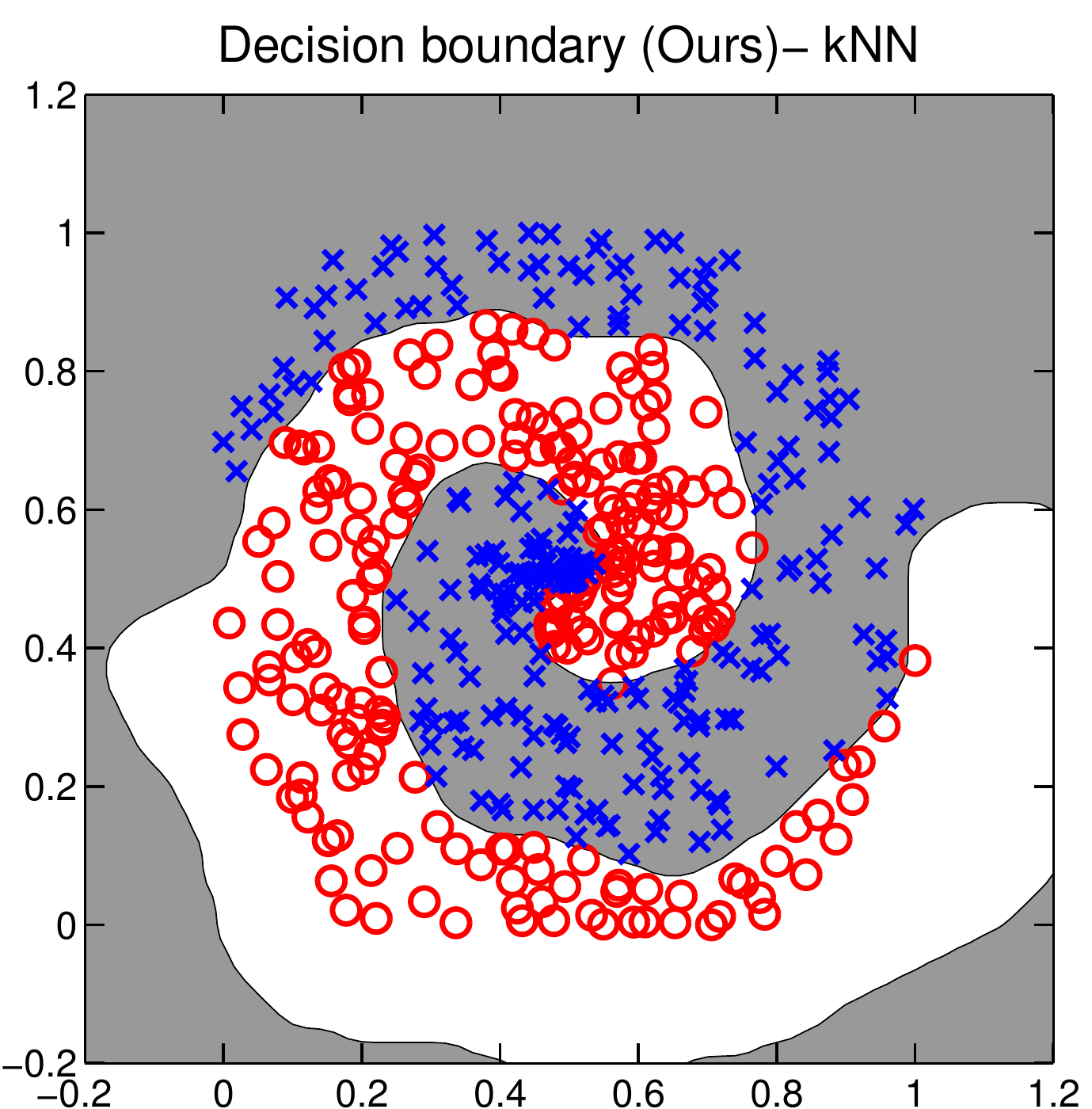}
        \includegraphics[width=0.24\textwidth,clip]{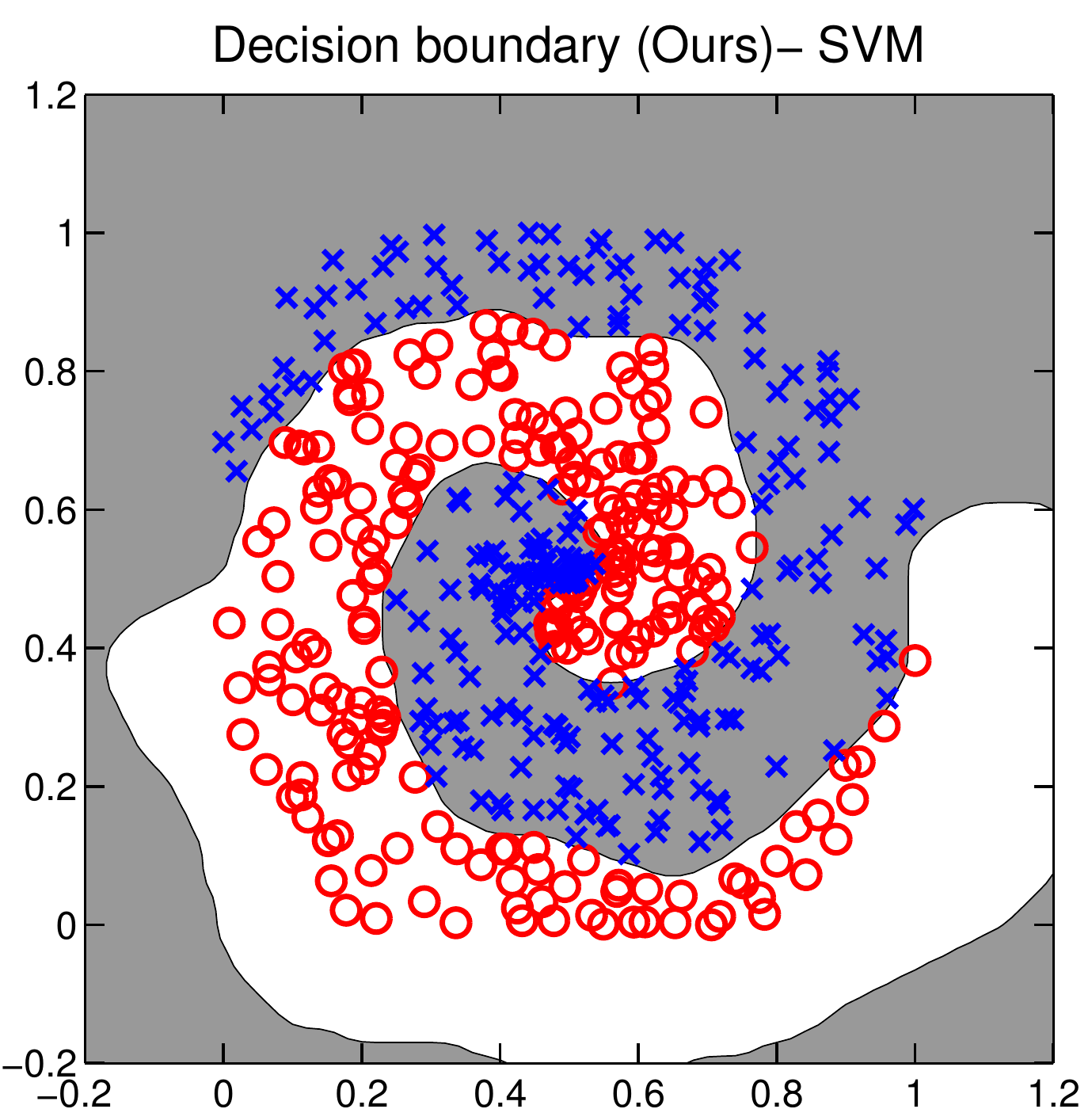}
    \end{center}
    \caption{
    {\em First column:} Decision boundary using kNN on a Fermat's problem
    ($k = 3$). {\em Second column:} Decision boundary using kNN on $8$
    bits code length learned with Spectral Hashing.
    {\em Last two columns:} Decision boundary using kNN and linear SVM
    on the $8$ bits feature learned using our approach.
    }
    \label{fig:toy}
\end{figure*}

\paragraph{Synthetic data set}
We first evaluate the behaviour of \algoname on a
Fermat's problem in which two-class samples
are distributed in a two-dimensional space forming a spiral shape.
We plot a decision boundary using kNN ($k = 3$) and illustrate
the result in Fig.~\ref{fig:toy}.
We then apply Spectral Hashing (SH) \cite{Weiss2008Spectral}
with $8$ bits code length
and plot the decision bounding using kNN in the same figure.
We observe that SH keeps
the neighbourhood relationship of the original data
but the decision boundary
is not as smooth as the decision boundary of
kNN on the original data.
We then evaluate our approach by training $8$ bits feature length
and plot the decision boundary using two popular classifiers
(kNN with $k=3$ and linear SVM).
The results are illustrated in the last two columns of Fig.~\ref{fig:toy}.
We observe that both classifiers achieve a very similar decision
boundary using our features.
Compared to the original $2$-D features, which do not exploit class
label information, the new feature has a much smoother decision boundary.
The results clearly demonstrates the benefit of training visual features
using both pair-wise proximity and class-label information.

\begin{figure}[tb]
    \begin{center}
        \includegraphics[width=0.05\textwidth,clip]{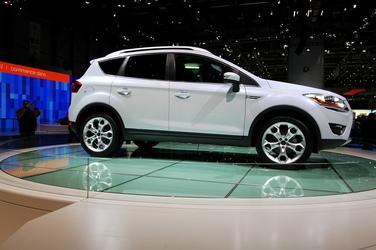}
        \includegraphics[width=0.05\textwidth,clip]{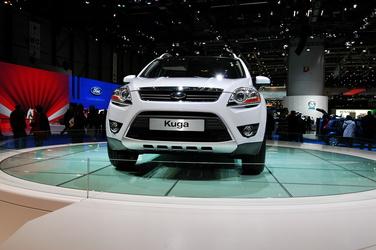}
        \includegraphics[width=0.05\textwidth,clip]{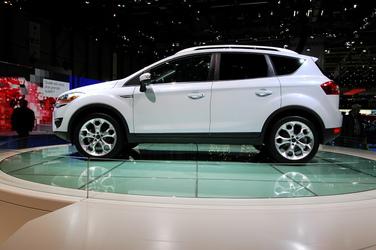}
        \includegraphics[width=0.05\textwidth,clip]{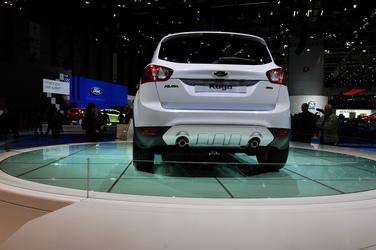}
        \hspace{.05cm}
        \includegraphics[width=0.05\textwidth,clip]{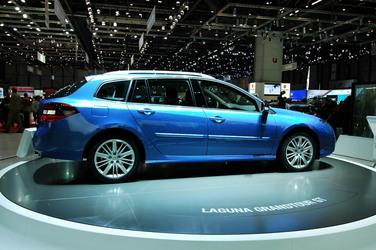}
        \includegraphics[width=0.05\textwidth,clip]{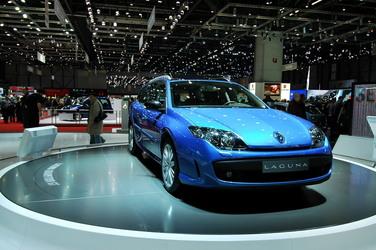}
        \includegraphics[width=0.05\textwidth,clip]{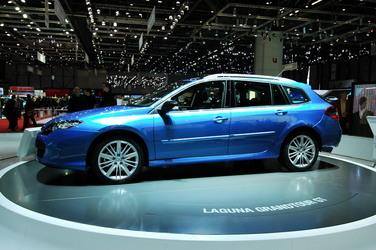}
        \includegraphics[width=0.05\textwidth,clip]{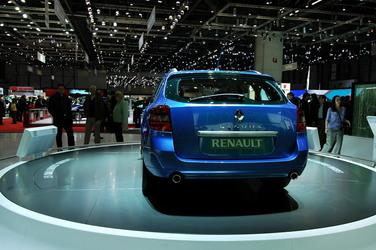} \\
        \includegraphics[width=0.05\textwidth,clip]{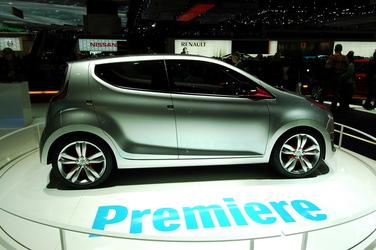}
        \includegraphics[width=0.05\textwidth,clip]{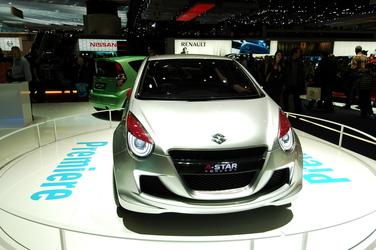}
        \includegraphics[width=0.05\textwidth,clip]{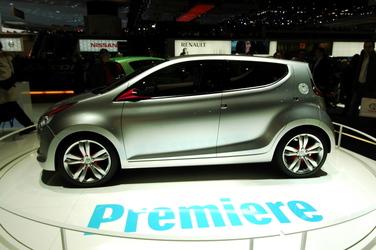}
        \includegraphics[width=0.05\textwidth,clip]{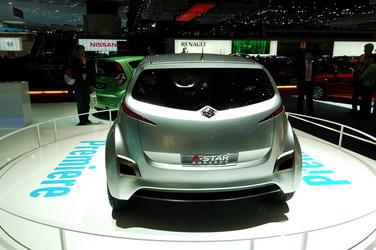}
        \hspace{.05cm}
        \includegraphics[width=0.05\textwidth,clip]{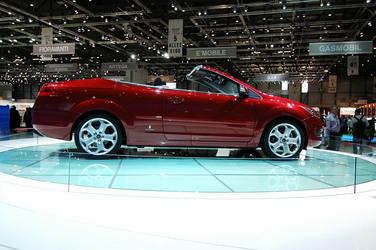}
        \includegraphics[width=0.05\textwidth,clip]{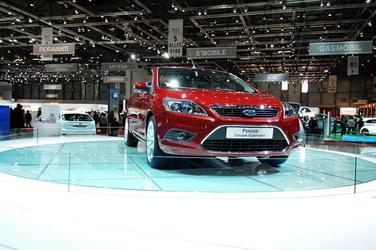}
        \includegraphics[width=0.05\textwidth,clip]{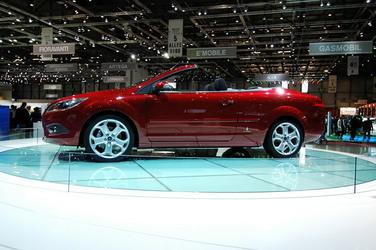}
        \includegraphics[width=0.05\textwidth,clip]{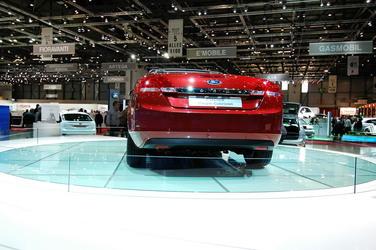} \\
        \includegraphics[width=0.45\textwidth,clip]{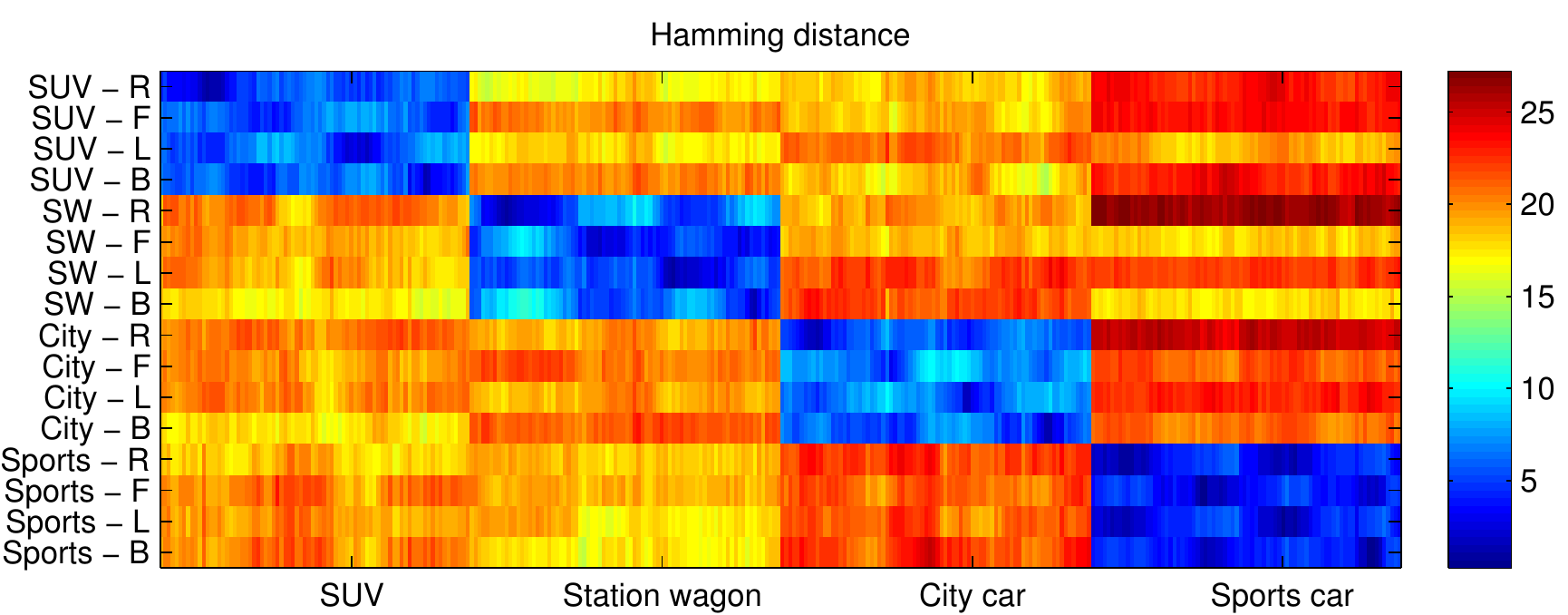}
    \end{center}
    \caption{
    {\em Top:} Each class corresponds to each vehicle type
    (SUV, station wagon (SW), city car and sports car) with
    orientations ranging from right-profile (R), frontal pose (F),
    left-profile (L) to back view (B).
    {\em Bottom:}
    The weighted hamming distance between the binary code of query
    images and that of the database is illustrated.
    Each row represents each query image and
    each column represents each vehicle type being rotated at $4-5$ degrees.
    Hamming distance can be separated into three groups:
    a set of training images having the same class and similar orientations
    to the query image (dark blue),
    a set of images having the same class but different orientations
    (light blue),
    and a set of images having different classes (orange).
    }
    \label{fig:EPFL}
\end{figure}

\paragraph{Image similarity identification}
In this experiment, we demonstrate the capability of \algoname
in retrieving similar images to the query image.
We use EPFL multi-view car data sets.
We categorize vehicles based on their appearance: SUV, station wagon,
city car and sports car.
Each category consists of the same vehicle type at different poses.
For each category, we randomly select $80$ images as the training set and
$4$ images, corresponding to the frontal pose, left-profile,
right-profile and back view, as the test set.
We extract pyramid HOG (pHOG) features and learn \algoname with $256$ bits.
The weighted hamming distance between the binary code of test images
and the binary code of training images
is plotted in Fig.~\ref{fig:EPFL}.
Each row (Fig.~\ref{fig:EPFL} Bottom) represents each query image, \eg,
SUV-R (the image of SUV captured at the right-profile orientation)
is the first query image.
Each column represents each vehicle type being rotated at $4-5$ degrees, \eg,
the first column represents the image of SUV
captured at the right-profile orientation and
the second column represents the image of SUV
rotated by $5$ degrees from the first column.
The binary code learned using \algoname is not only
discriminative across different vehicle types but also preserves similarity
within the same vehicle category.

\begin{figure}[tb]
    \begin{center}
        \includegraphics[width=0.23\textwidth,clip]{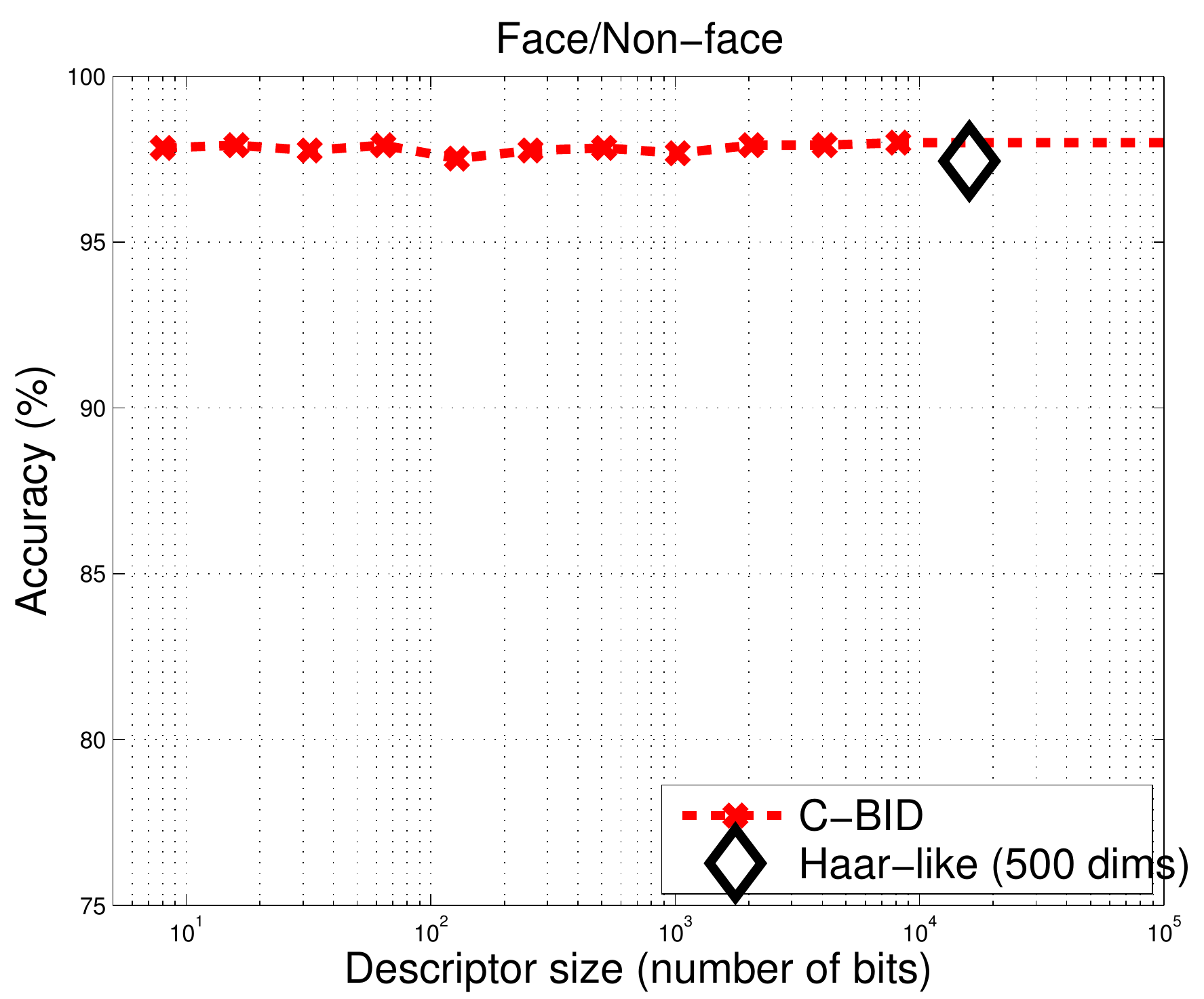}
        \includegraphics[width=0.23\textwidth,clip]{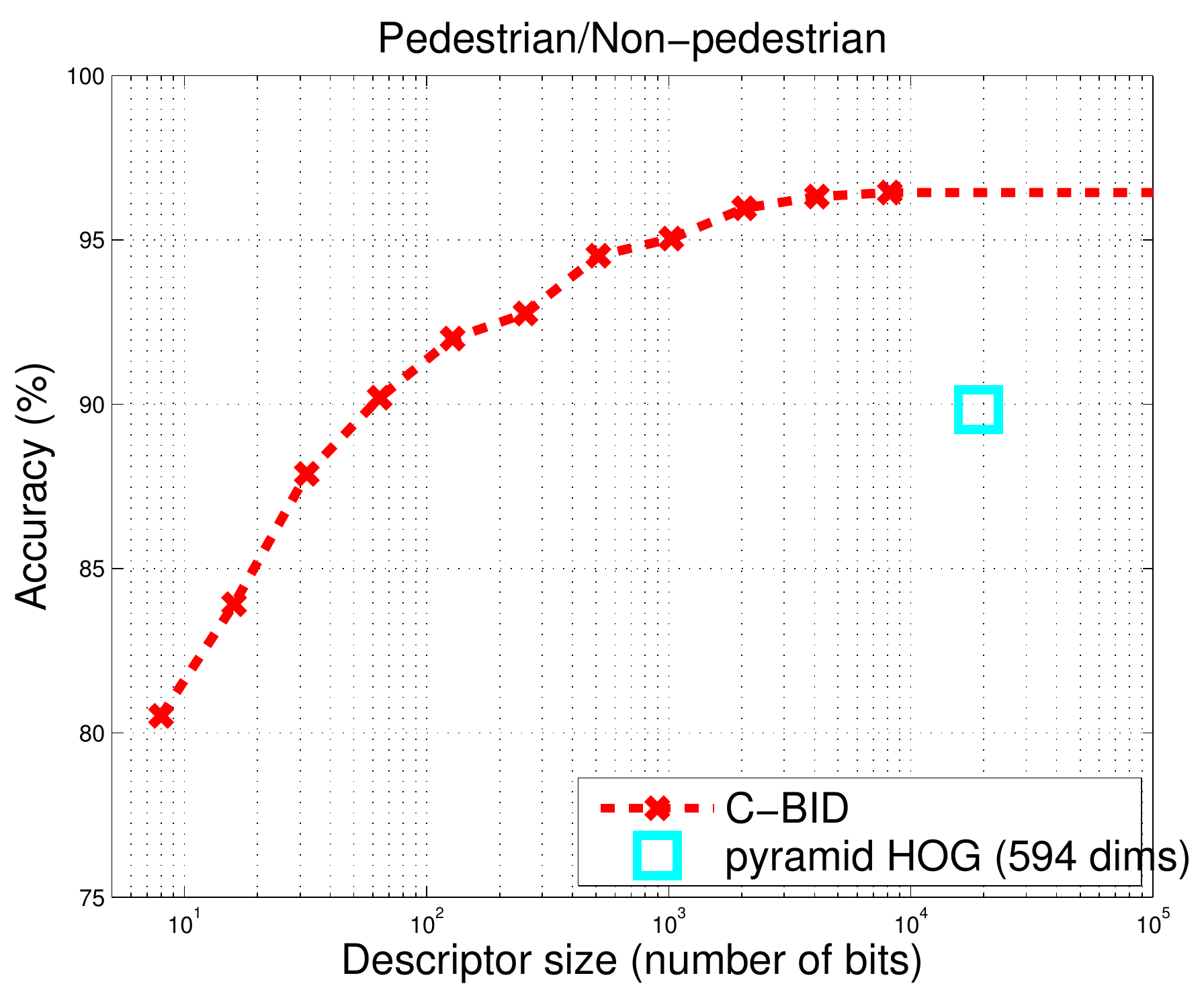}
    \end{center}
    \caption{
    The average face ({\em left}) and pedestrian ({\em right})
    classification performance on various code lengths.
    }
    \label{fig:exp_det_patch}
\end{figure}

\paragraph{Face and pedestrian classification}
Next we evaluate the performance of \algoname on face and pedestrian
classification.
We use $2000$ faces from \cite{Viola2004Robust} and randomly extract
$2000$ negative patches from background images.
We use the original raw pixel intensity and Haar-like features in our experiments.
For pedestrians, we use the Daimler-Chrysler pedestrian data sets \cite{Munder2006Experimental}.
We use the original raw pixel intensities and pHOG features.
We use $5$ nearest neighbours information per training sample.
On both face and pedestrian data sets,
\algoname consistently outperforms raw intensity, Haar-like and
pHOG features.
We show the results of experiments using both Haar-like and pHOG features in Fig.~\ref{fig:exp_det_patch}.

\begin{figure}[tb]
    \begin{center}
        \includegraphics[width=0.23\textwidth,clip]{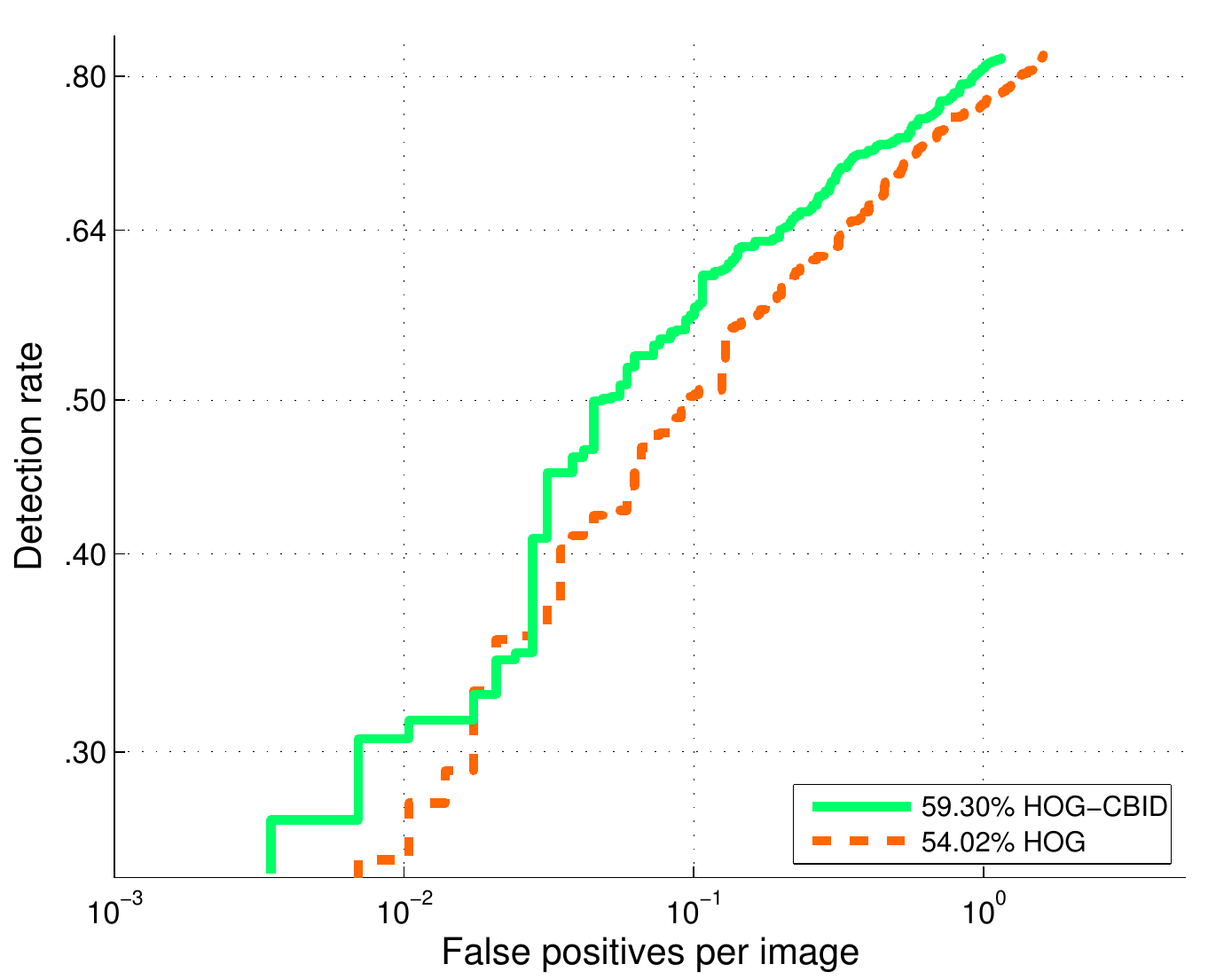}
        \includegraphics[width=0.23\textwidth,clip]{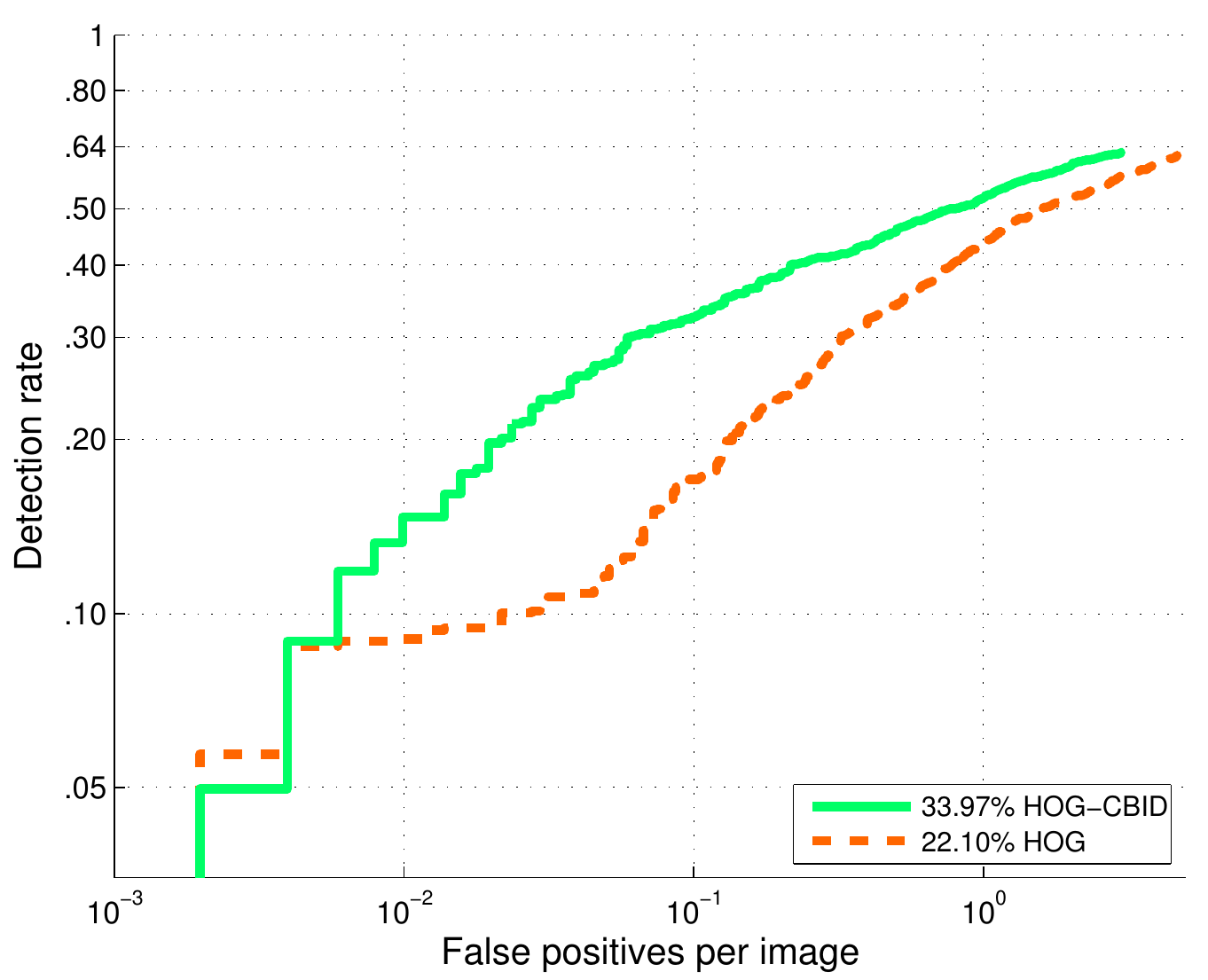}
    \end{center}
    \caption{
    Pedestrian detection performance ({\em left}) INRIA
    ({\em right}) TUD-Brussels.
    Log-average detection rate is also reported.
    }
    \label{fig:human_det}
\end{figure}

\begin{figure}[tb]
    \begin{center}
        \includegraphics[width=0.23\textwidth,clip]{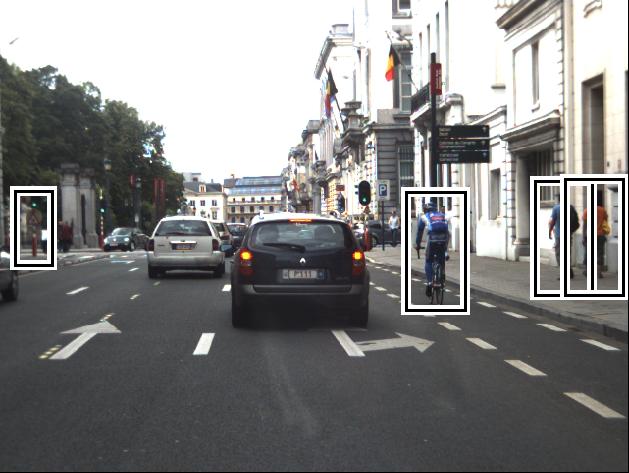}
        \includegraphics[width=0.23\textwidth,clip]{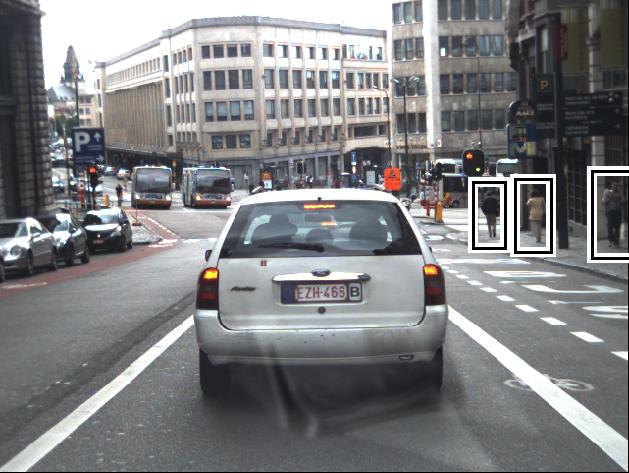}
        \includegraphics[width=0.23\textwidth,clip]{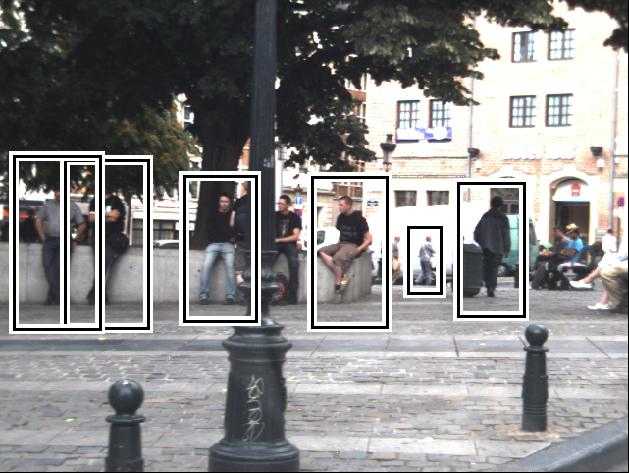}
        \includegraphics[width=0.23\textwidth,clip]{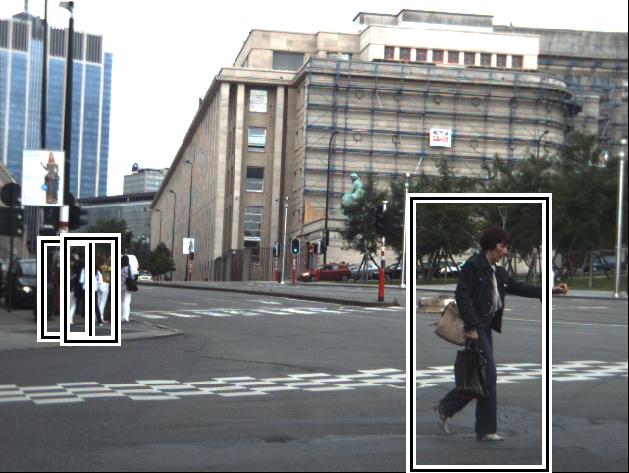}
        \includegraphics[width=0.23\textwidth,clip]{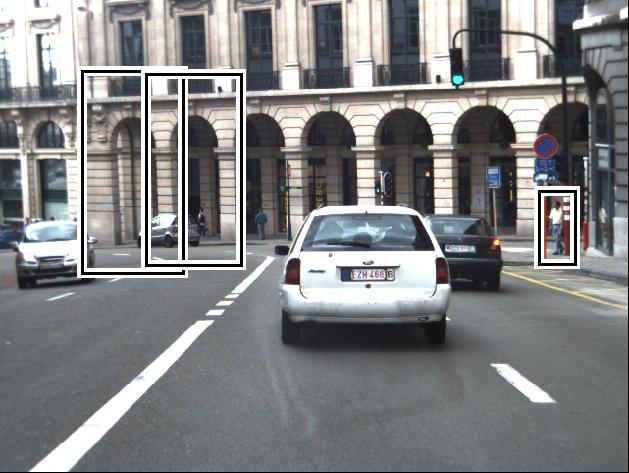}
        \includegraphics[width=0.23\textwidth,clip]{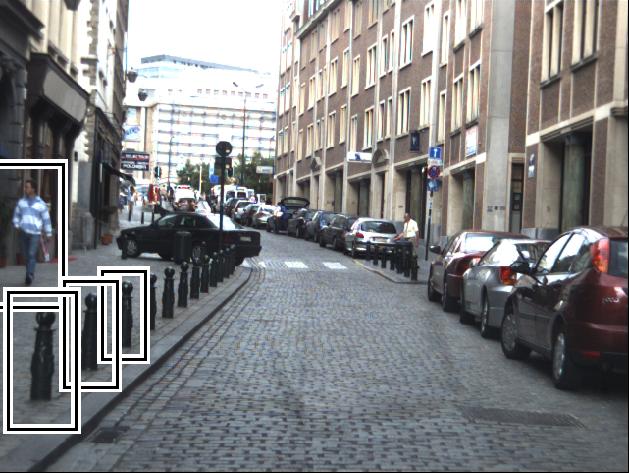}
    \end{center}
    \caption{
    {\em Top two rows:} Pedestrian detection examples on TUD-Brussels data sets.
    {\em Last row:} False positive examples made by our detector.
    Most false positive examples usually contain
    patterns that mimic the contour of human shoulders
    or vertical gradients that mimic the torso and leg boundaries.
    }
    \label{fig:human_det_examples}
\end{figure}

In the next experiment, we apply \algoname to the baseline pedestrian detector
of \cite{Dalal2005Histograms} using INRIA human data sets \cite{Dalal2005Histograms}
and TUD-Brussels data sets \cite{Wojek2009Multi}.
INRIA training set consists of $2,416$ cropped mirrored pedestrian images
and $1,200$ large background images.
The test set contains $288$ images containing $588$ annotated pedestrians and $453$
non-pedestrian images.
In this experiment, we only use $288$ images which contain
pedestrians \cite{Dollar2012Pedestrian}.
TUD-Brussels test set contains $508$ images containing $1498$ annotated pedestrians.
Pedestrian detection is divided into two steps: initial classification
and post-verification.
The first step is similar to the method described in \cite{Dalal2005Histograms}
while, in the second step, we apply \algoname to enforce the pairwise-similarity
between positive training samples.
To be more specific, in the first step,
each training sample is scaled to $64 \times 128$ pixels with $16$ pixels
additional borders for preserving the contour information.
Histogram of oriented gradient features are extracted from $105$ blocks of
size $16 \times 16$ pixels.
Each block is divided into $2 \times 2$ cells, and the HOG in each cell
is divided into $9$ bins.
We train a pedestrian detector using the linear SVM.
In the second step, we learn \algoname from HOG extracted in the first step.
We use $5$ nearest neighbours information to generate $25$ triplets
per positive training image.
We learn $3780$ bits descriptor (similar to the original HOG feature length)
and build the classifier using the linear SVM.
During evaluation, each test image is scanned with $4 \times 4$ pixels
step size and $10$ scales per octave.
For TUD-Brussels, we up-sample the original image to $1280 \times 960$ pixels
before applying the pedestrian detector.
The performance is evaluated using the protocol described in
\cite{Dollar2012Pedestrian}.
Both ROC curves are plotted in Fig.~\ref{fig:human_det}.
We also report the $\log$-average detection rate which is
computed by averaging the detection rate
at nine FPPI rates evenly spaced in $\log$-space
between $10^{-2}$ to $1$.
We observe that applying \algoname further improves the
$\log$-average detection rate by $5.28\%$ on INRIA test set
and $11.87\%$ on TUD-Brussels data sets.
This performance gain comes at no additional feature extraction cost.
Fig.~\ref{fig:human_det_examples} shows a qualitative assessment of the new detector.
We observe that most false positive examples usually contain
patterns that mimic the contour of human shoulders
or vertical gradients that mimic the torso and leg boundaries.

\begin{figure}[tb]
    \begin{center} %
        \includegraphics[width=0.23\textwidth,clip]{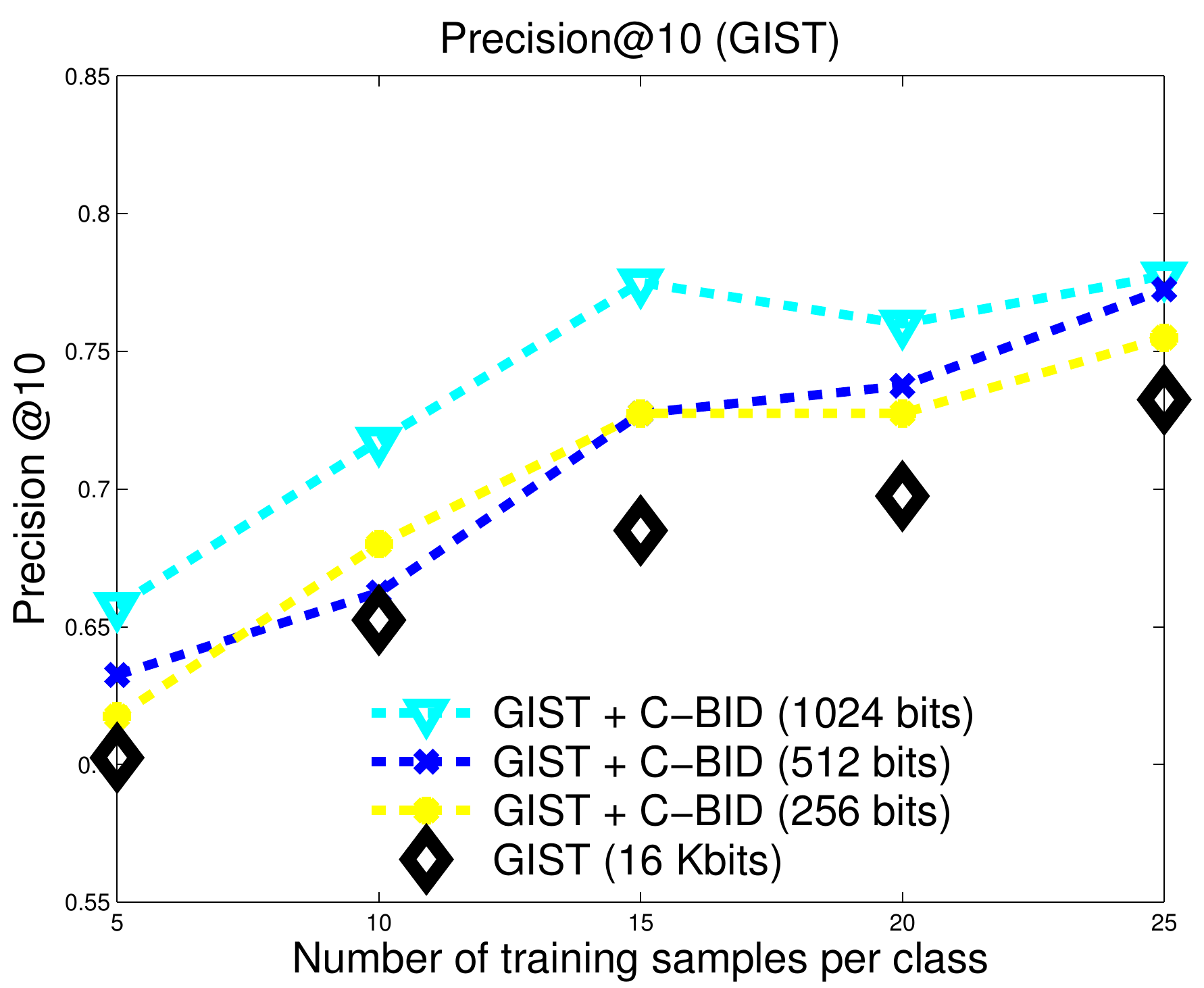}
        \includegraphics[width=0.23\textwidth,clip]{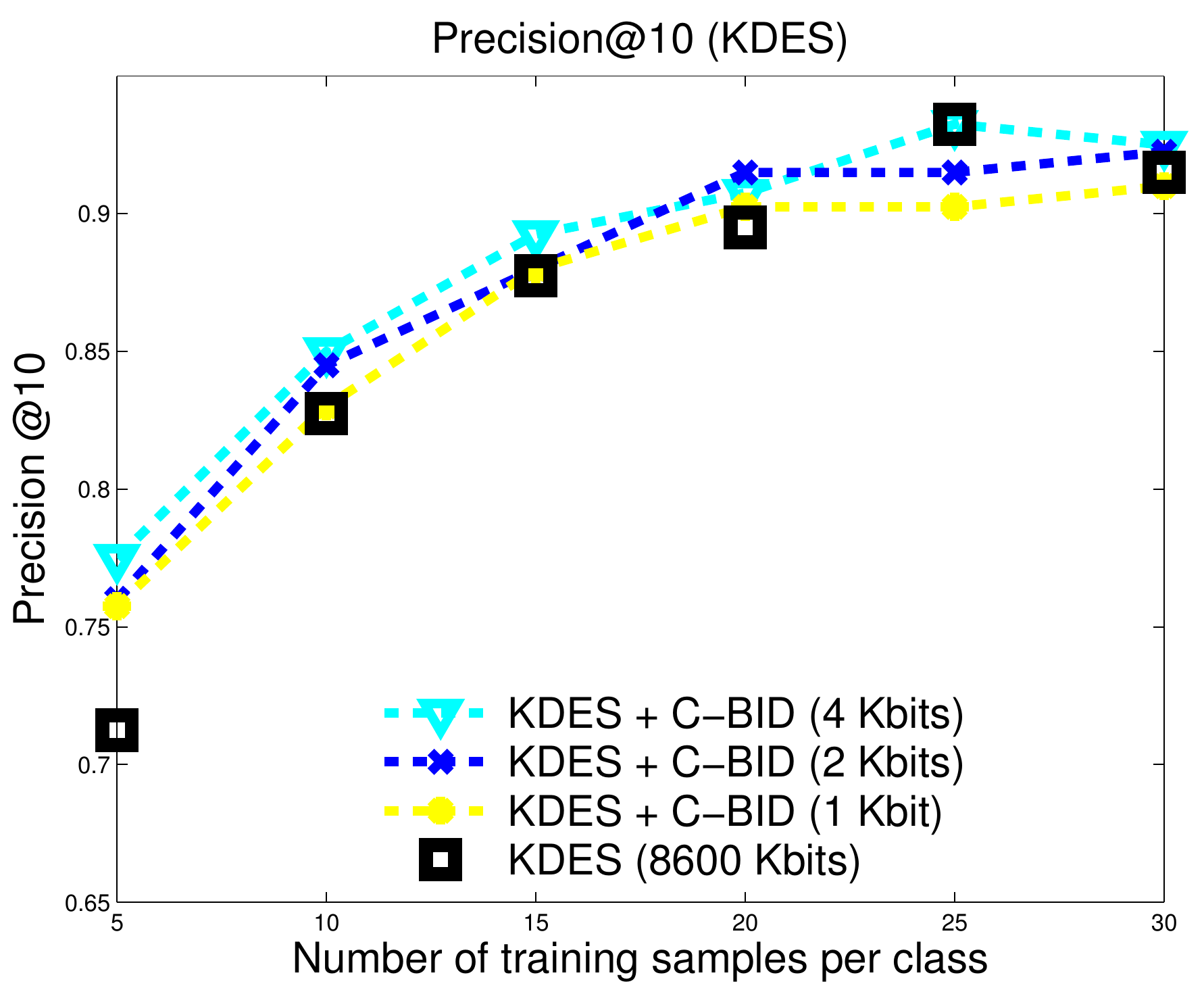}
        \includegraphics[width=0.23\textwidth,clip]{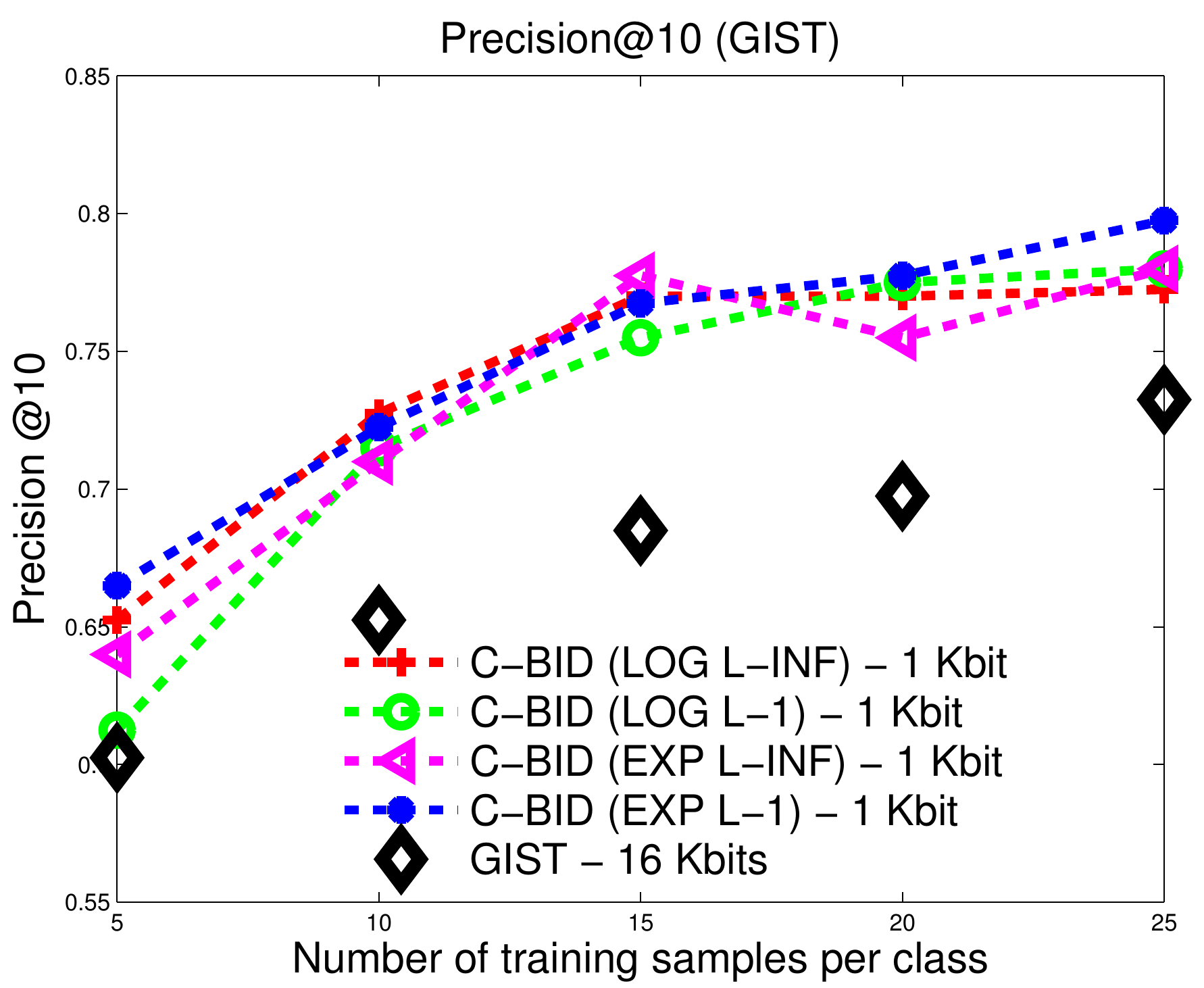}
        \includegraphics[width=0.23\textwidth,clip]{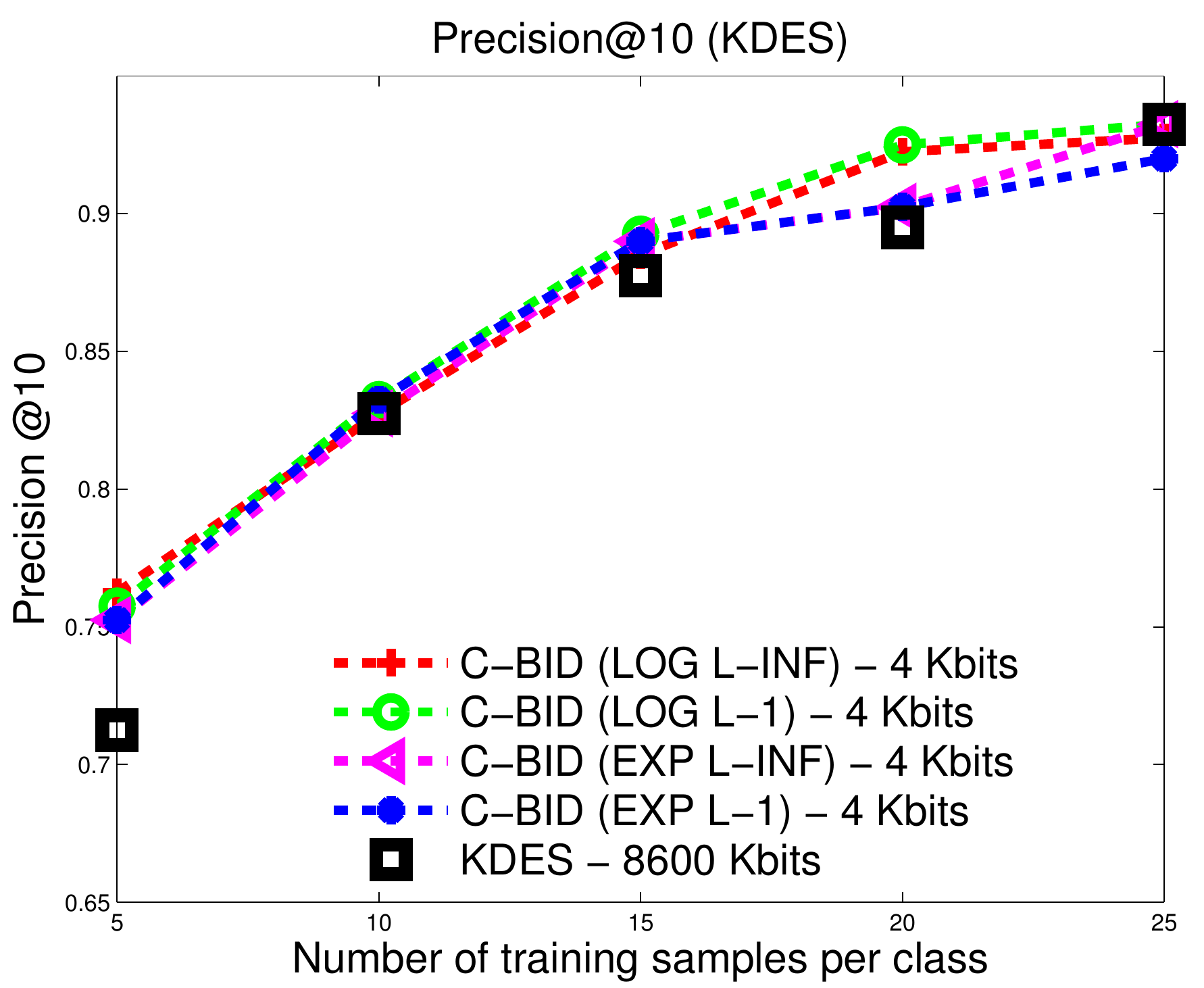}
    \end{center}
    \caption{
    Combining \algoname with GIST and KDES.
    {\em Top:} Average `precision rate at $10$' of different
    number of training samples per class.
    {\em Bottom:} Average precision rate of logistic and exponential losses
    with $\ell_1$ and $\ell_\infty$ penalties.
    Using GIST, \algoname at $256$ bits consistently outperforms
    GIST ($512$ dims $\approx$ $16$ kilobits).
    Using KDES, \algoname at $4096$ bits outperforms
    KDES ($270,000$ dims $\approx$ $8.6$ megabits).
    }
    \label{fig:exp_i2i_gist}
\end{figure}

\paragraph{Combining \algoname with various visual descriptors}
Next we evaluate the performance of \algoname using
GIST \cite{Torralba2003Context} and recently proposed
Kernel Descriptors (KDES) \cite{Bo2011Object}.
We use UIUC Sports-8 %
data sets and vary the
number of training samples per class.
For each split, we train a linear SVM.
The results are reported in Fig.~\ref{fig:exp_i2i_gist}.
We observe that \algoname at $256$ bits consistently outperforms GIST
(GIST occupies $4 \times 512$ bytes of storage space)
and \algoname at $2048$ bits performs comparable to KDES.

In the next experiment, we compare the performance of our binary descriptors
with several existing bit-code-based methods.
We use the state-of-the-art feature descriptor known as
Meta-Class features (MC) \cite{Bergamo2012MetaClass} to learn binary codes.
Experiments are conducted on four different visual data sets:
Graz-02, %
UIUC Sports-8, %
Scene-15, %
and
Caltech-101. %
For Graz-02 and Sports-8, we randomly select $50$ images per class for training,
$10$ images per class for validation and the remainder for testing.
For Scene-15, we randomly select $100$ images for training, $10$ images for validation,
and the remainder for testing.
For Caltech-101, we randomly select $15$ images per class for training,
$5$ images per class for validation, and $20$ images per class for test.
We evaluate our algorithm using two different classifiers: K-NN and SVM.

\begin{figure*}[t]
    \begin{center}
        \includegraphics[width=0.24\textwidth,clip]{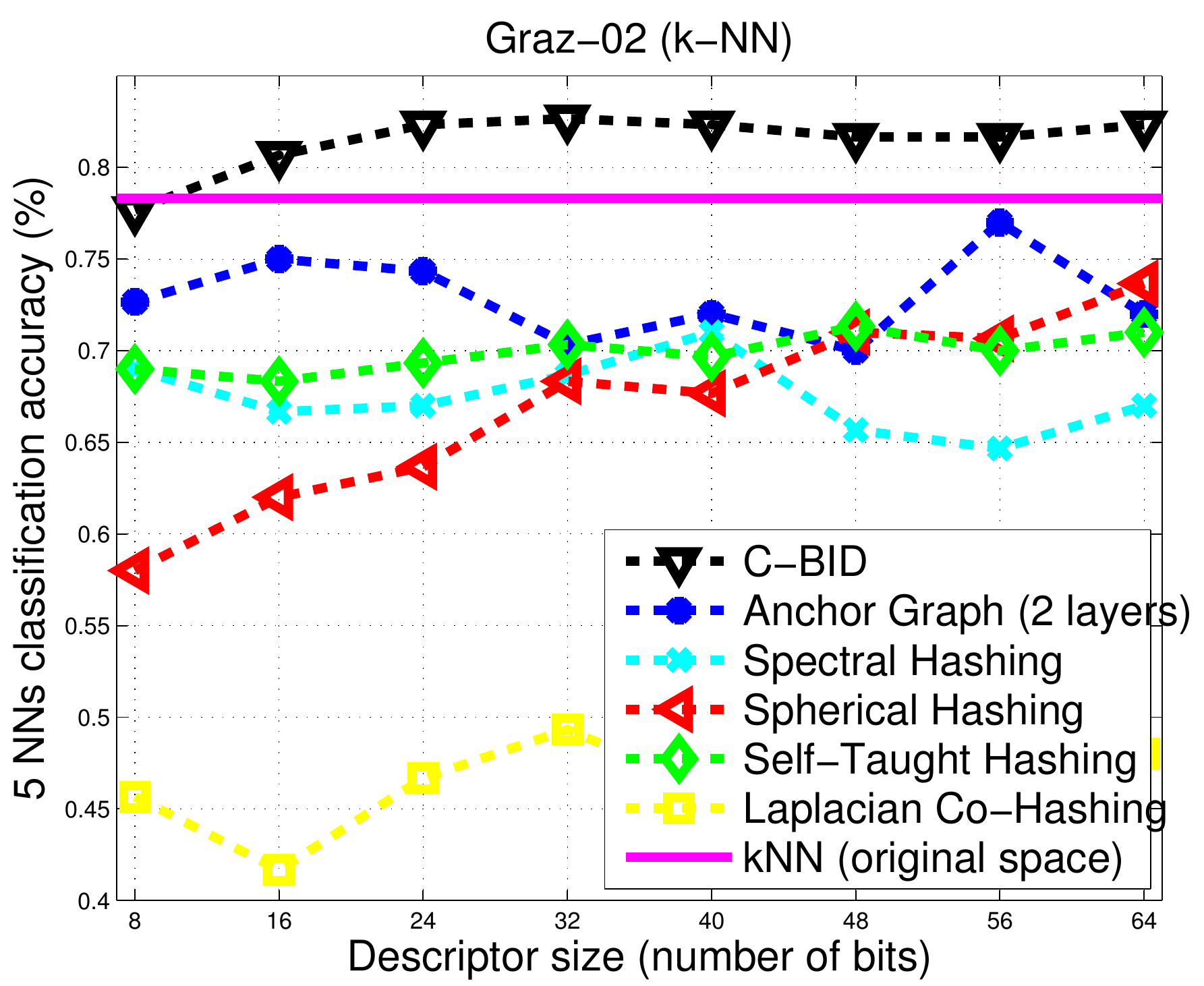}
        \includegraphics[width=0.24\textwidth,clip]{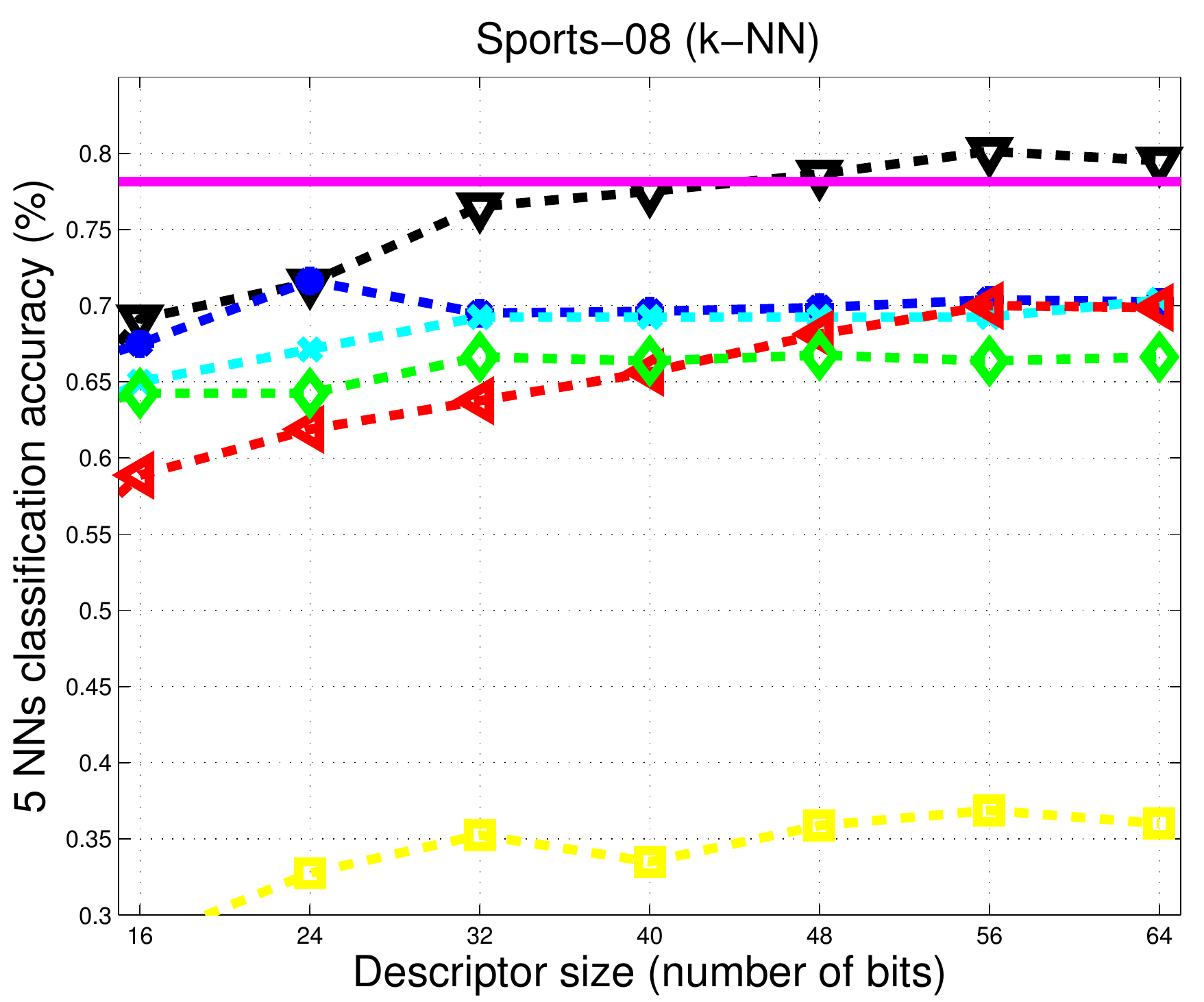}
        \includegraphics[width=0.24\textwidth,clip]{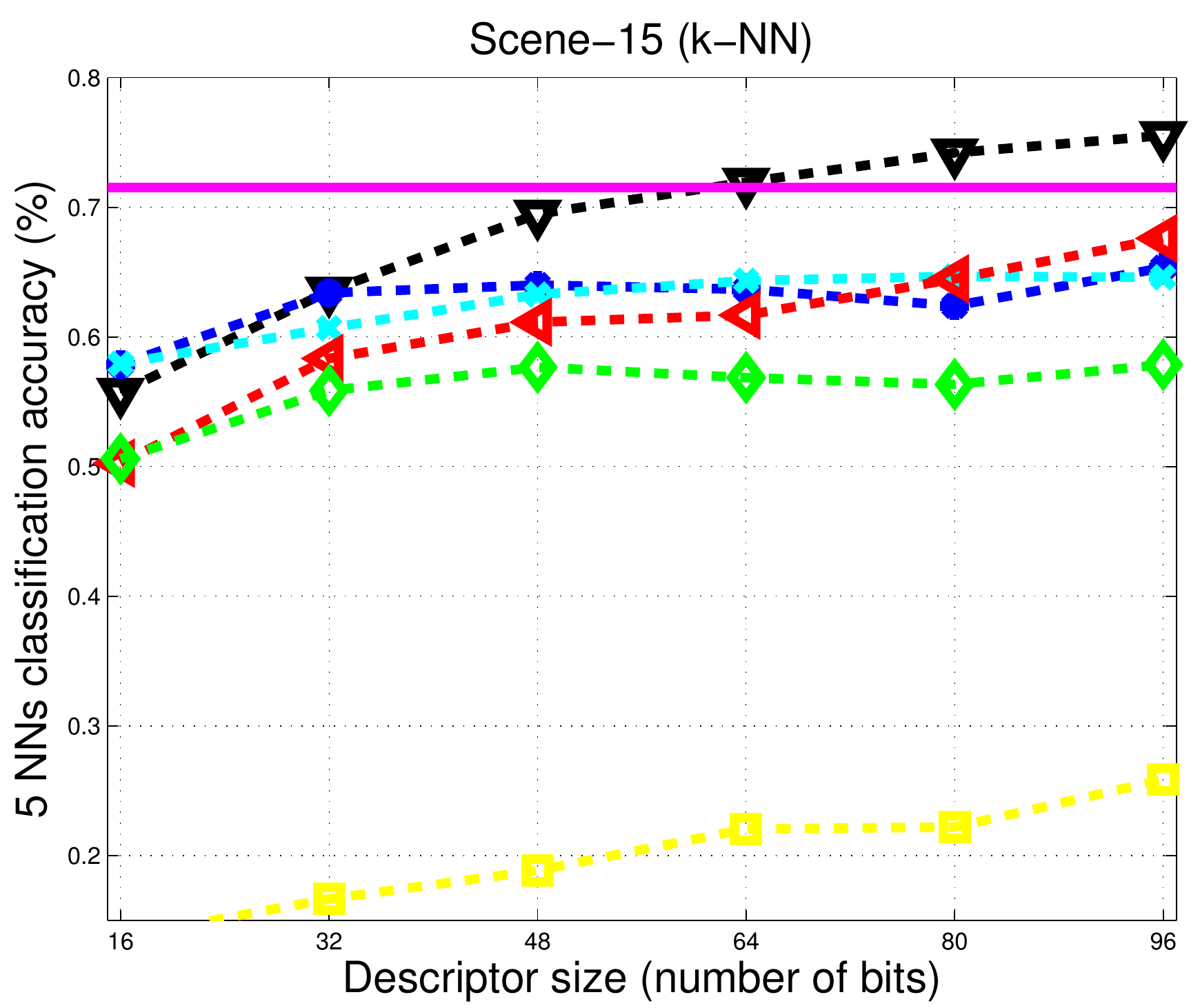}
        \includegraphics[width=0.24\textwidth,clip]{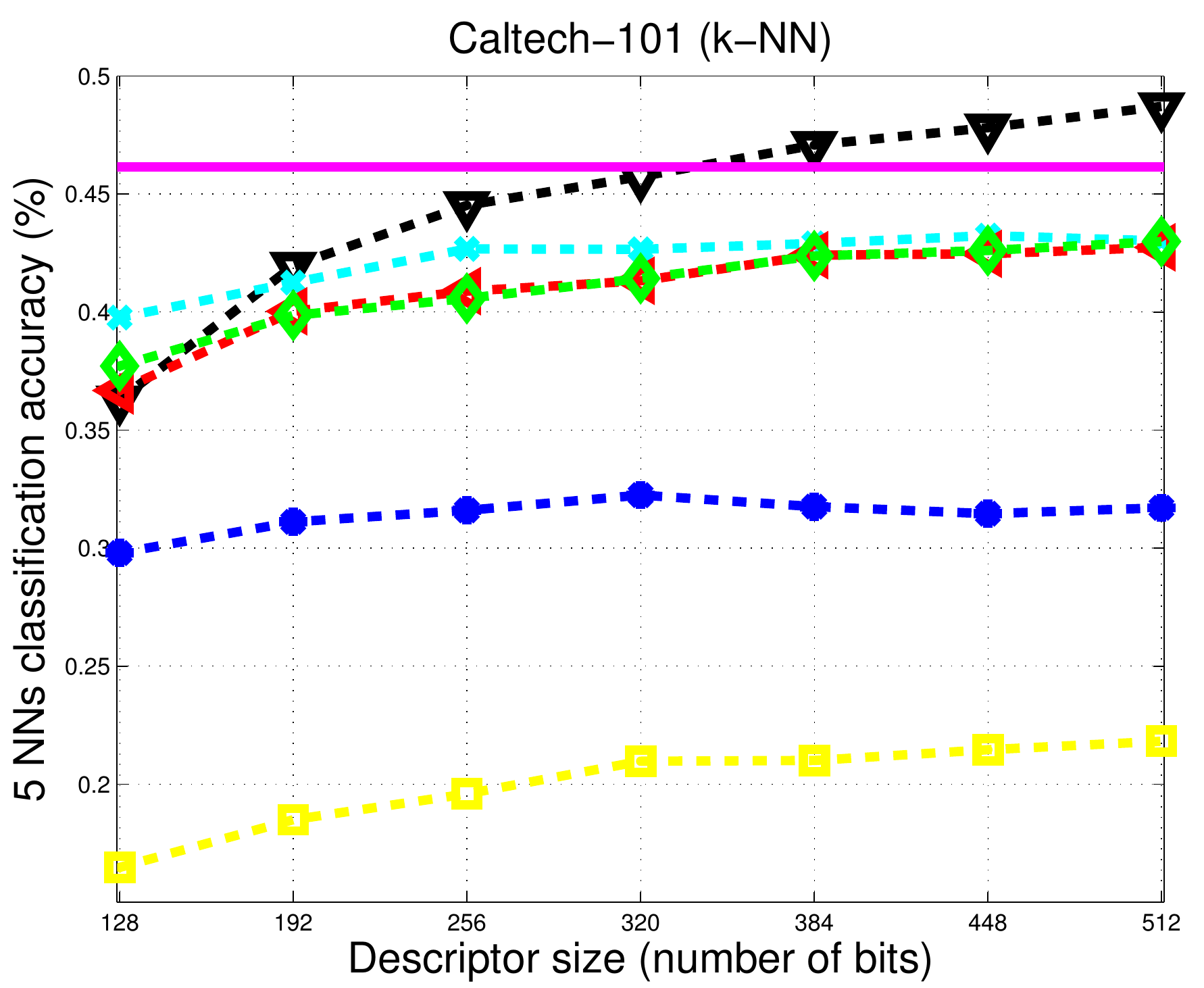}
    \end{center}
    \caption{
    The average k-NN classification performance.
    We compare our approach with Anchor Graph \cite{Liu2011Hashing},
    Spectral Hashing \cite{Weiss2008Spectral},
    Spherical Hashing \cite{Heo2012Spherical},
    Self-Taught Hashing \cite{Zhang2010Self}, and
    Laplacian Co-Hashing \cite{Zhang2010Laplacian}.
    The performance of kNN is also plotted.
    \algoname outperforms all hashing methods
    as the number of bits increases.
    }
    \label{fig:exp_i2i1}
\end{figure*}

\begin{figure*}[t]
    \begin{center}
        \includegraphics[width=0.24\textwidth,clip]{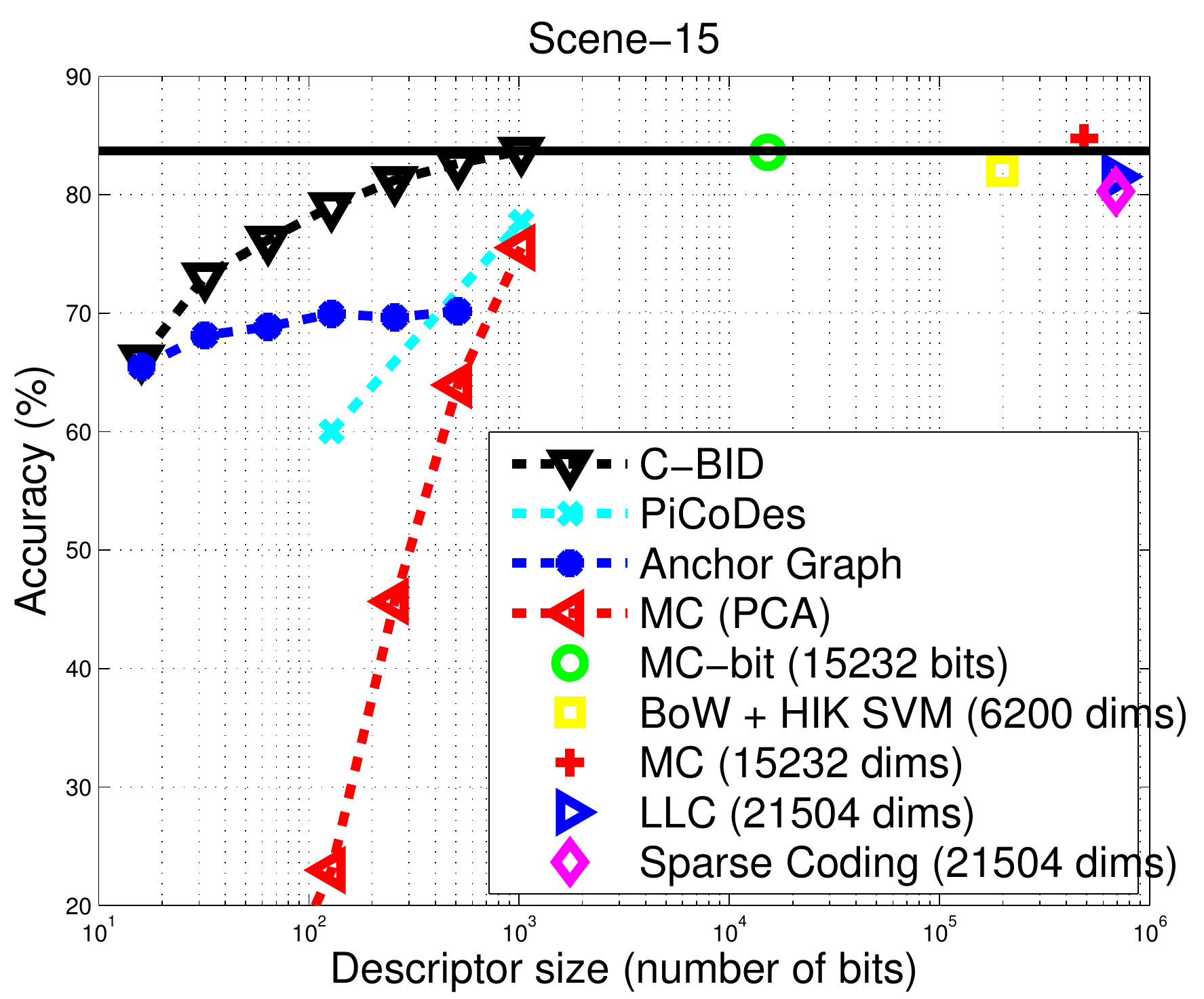}
        \includegraphics[width=0.24\textwidth,clip]{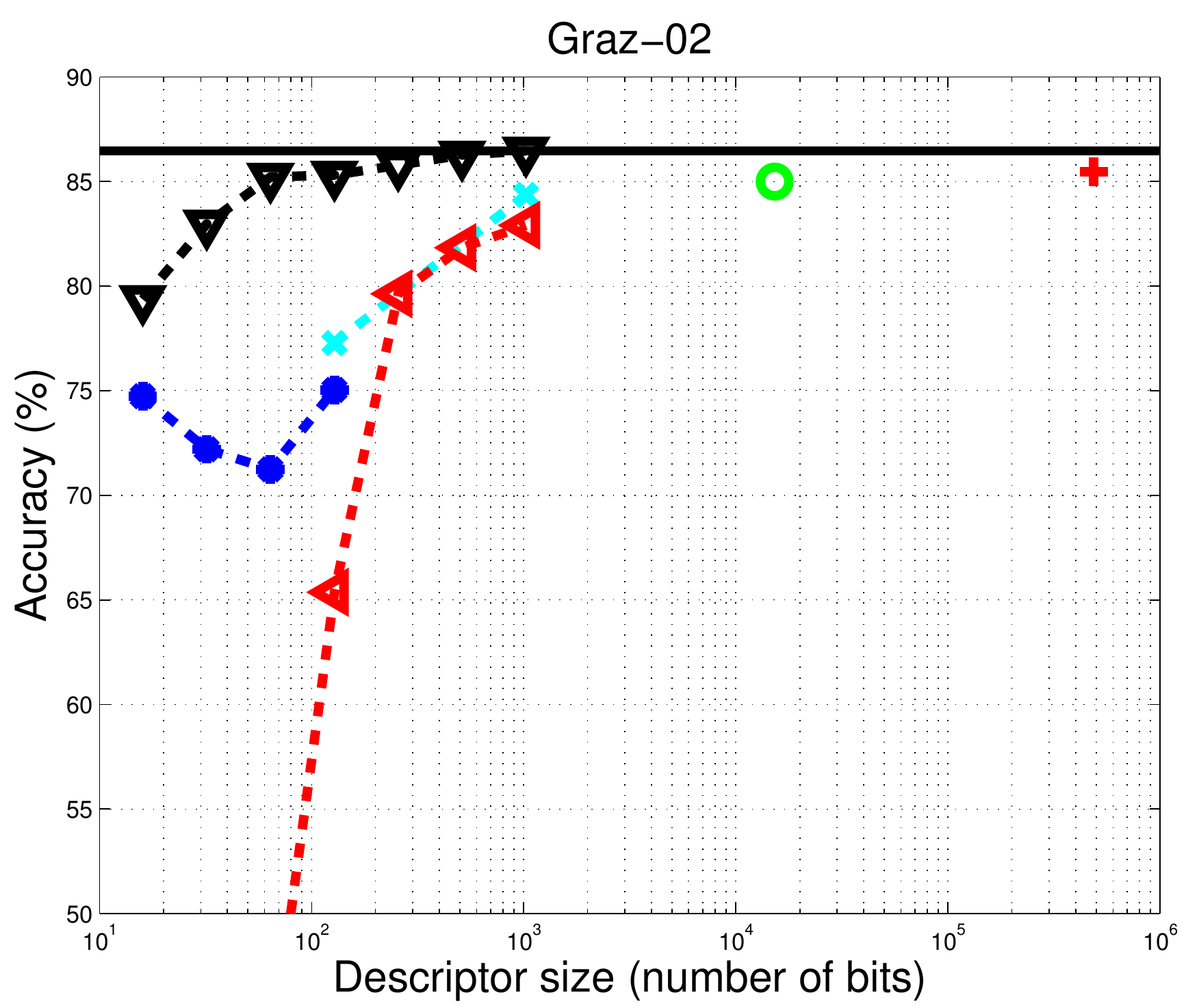}
        \includegraphics[width=0.24\textwidth,clip]{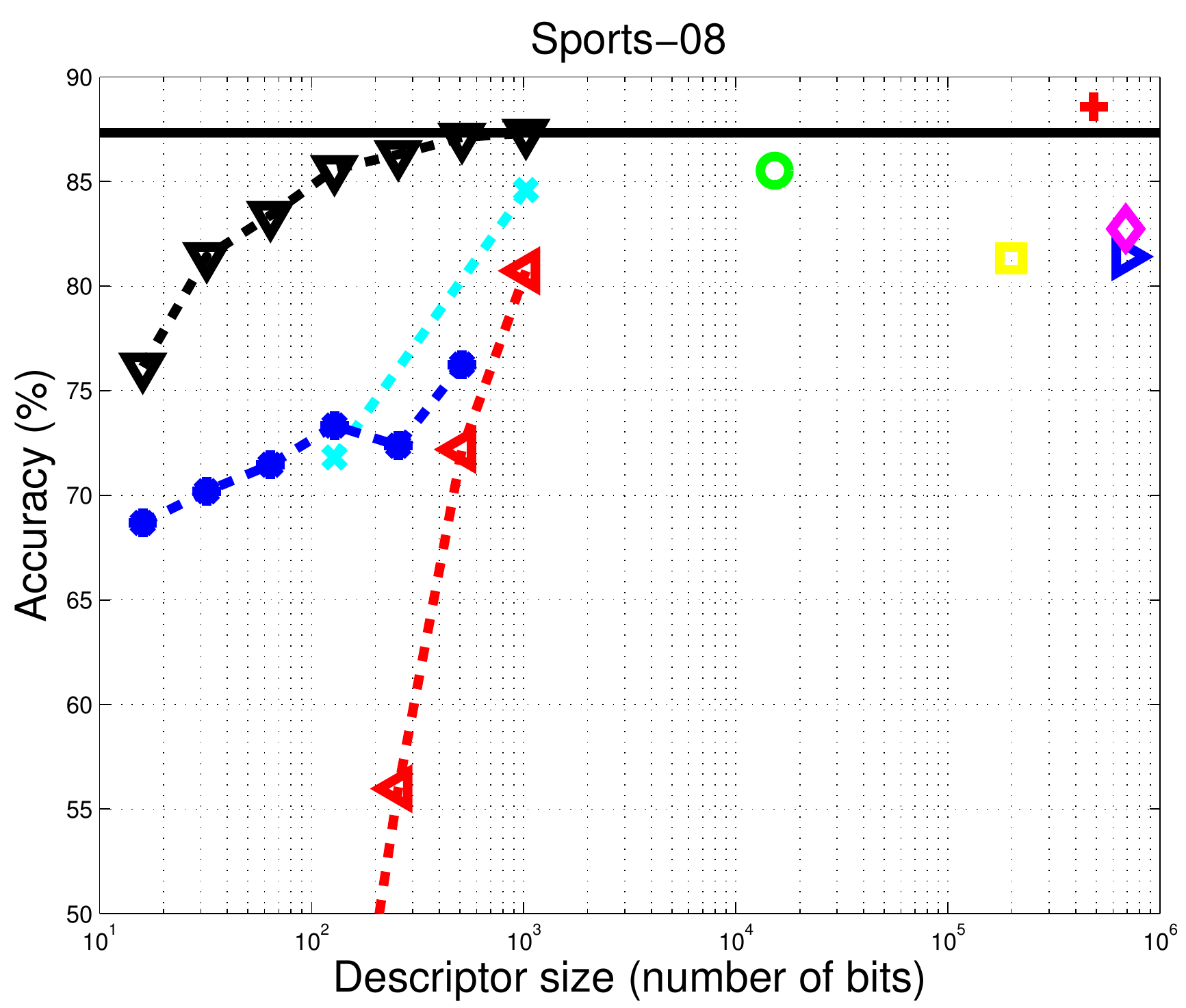}
        \includegraphics[width=0.24\textwidth,clip]{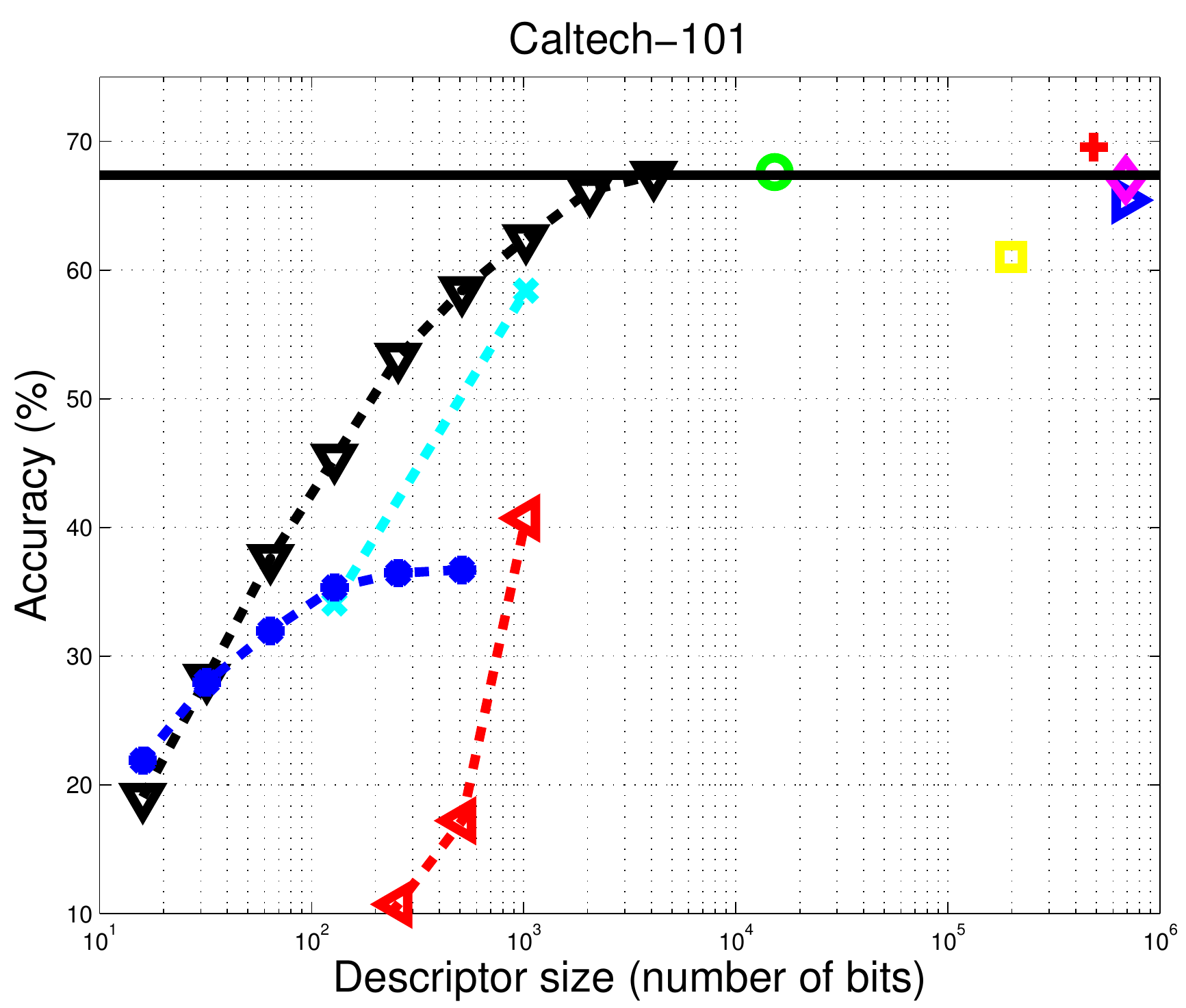}
    \end{center}
    \caption{
    Image classification accuracy on Scene-15, Graz-02, Sports-8 and
    Caltech-101 data sets using different binary codes as a function of
    the number of bits.
    We compare our results with
    AGHash \cite{Liu2011Hashing} + Linear SVM,
    MC-bit \cite{Bergamo2012MetaClass},
    BoW + HIK SVM \cite{Wu2011CENTRIST},
    Sparse Coding \cite{Yang2009Linear} and
    LLC \cite{Wang2010Locality}.
    \algoname achieves comparable accuracy to many state-of-the-art algorithms
    while requiring {\em at least an order of magnitude less storage}.
    }
    \label{fig:exp_i2i2}
\end{figure*}

\paragraph{k-NN}
In this experiment, we compare \algoname with various hashing algorithms using k-NN
classification accuracy.
For \algoname, we use $5$ nearest neighbours information
to generate $25$ triplets per training image.
We use MC-PCA as our feature descriptors ($99\%$ of the total variance is preserved).
We compare our approach with various state-of-the-art hashing methods, \eg
Spectral Hashing (SH) \cite{Weiss2008Spectral},
Two-layer Anchor Graph Hashing (AGH) \cite{Liu2011Hashing},
Spherical Hashing (SphH) \cite{Heo2012Spherical},
Self-Taught Hashing (STH) \cite{Zhang2010Self} and
Laplacian Co-Hashing (LCH) \cite{Zhang2010Laplacian}.
For AGHash, we set the number of anchors to be $300$ and
the number of nearest anchors to be $5$.
For STH, we set the k-NN parameter for LapEig to be $5$.
The experiments are repeated $5$ times and the average accuracy is reported in Fig.~\ref{fig:exp_i2i1}.
On Scene-15 data sets, we achieve the same accuracy as the original
features while requiring only $64$ bits.
By increasing the number of bits, our approach outperforms all hashing
algorithms.
Clearly, the performance of our approach improves as the number of
bits increases.
Examples of retrieval results for Graz and Sports data sets are shown in
Table~\ref{tab:graz1} and \ref{tab:sports1}.

\paragraph{SVM}
To compare \algoname with state-of-the-art methods,
\eg, Sparse Coding \cite{Yang2009Linear} and
LLC \cite{Wang2010Locality}, we learn the classifier using SVM.
We also compare \algoname with other binary descriptors,
\eg, \picodes \cite{Bergamo2011PICODES},
the thresholded version of MC (known as MC-bit \cite{Bergamo2012MetaClass}),
MC-PCA and AGHash.
For \algoname, we choose the regularization parameter from
$\{ 10^{-8}, 10^{-7}, 10^{-6}, 10^{-5}\}$.
We use $5$ nearest neighbours information
to generate $25$ triplets per training image.
Careful tuning of these parameters can further yield an
improvement from the results reported here.
We use LIBLINEAR \cite{Fan2008Liblinear} to train \algoname, \picodes, AGHash
and MC descriptors.
Average classification accuracy for different algorithms is
compared in Fig.~\ref{fig:exp_i2i2}.
We observe that our approach outperforms the baseline MC-PCA descriptors and
performs comparable to many state-of-the-art algorithms when the number of bits increases.
We observe that AGHash+SVM performs quite poorly as the number of bits increases.
The similar phenomenon has also been pointed out in \cite{Weiss2012Multi}.

\begin{table*}[tb]
\caption{Examples of top $10$ retrieval results for \textbf{bicycle} (Graz data sets). Red bounding box indicates retrieval error}
\begin{tabular}{ccc}
  & \algoname & \adjustbox{valign=m}{\includegraphics[width=0.8\textwidth,clip]{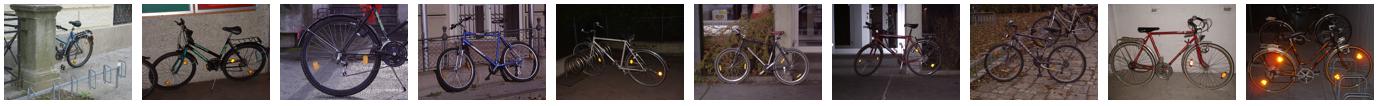}} \\
  & AGH \cite{Liu2011Hashing} & \adjustbox{valign=m}{\includegraphics[width=0.8\textwidth,clip]{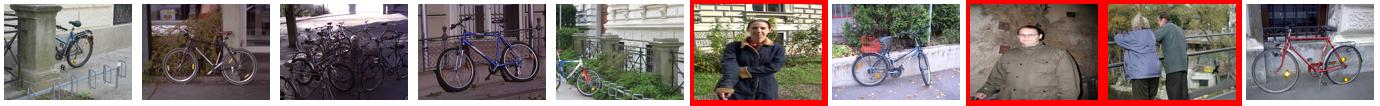}} \\
  \multirow{7}{*}{\includegraphics[width=0.1\textwidth,clip]{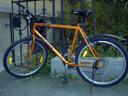}}   & LCH \cite{Zhang2010Laplacian} & \adjustbox{valign=m}{\includegraphics[width=0.8\textwidth,clip]{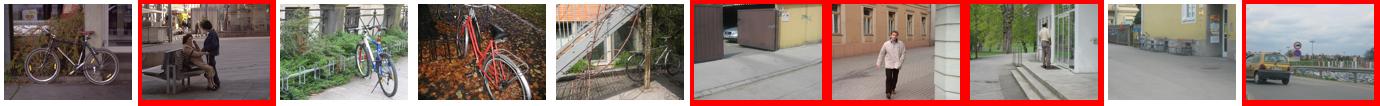}} \\
  & LSI \cite{Deerwester1990Indexing} & \adjustbox{valign=m}{\includegraphics[width=0.8\textwidth,clip]{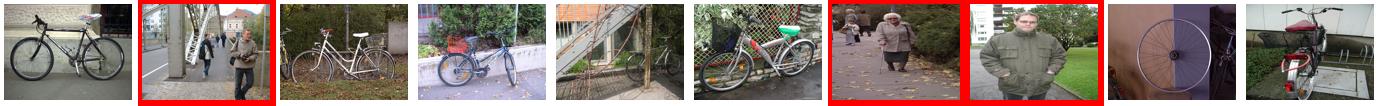}} \\
  & SH \cite{Weiss2008Spectral} & \adjustbox{valign=m}{\includegraphics[width=0.8\textwidth,clip]{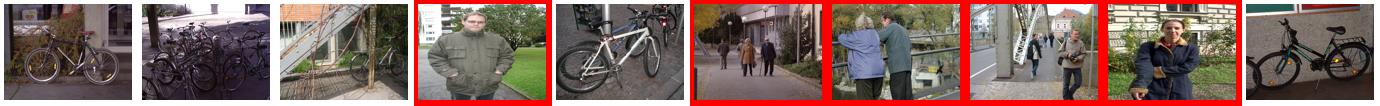}} \\
  & SphH \cite{Heo2012Spherical} & \adjustbox{valign=m}{\includegraphics[width=0.8\textwidth,clip]{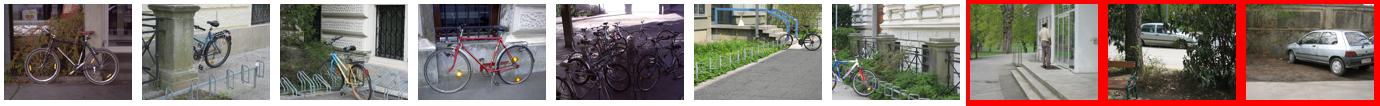}} \\
   & STH \cite{Zhang2010Self} & \adjustbox{valign=m}{\includegraphics[width=0.8\textwidth,clip]{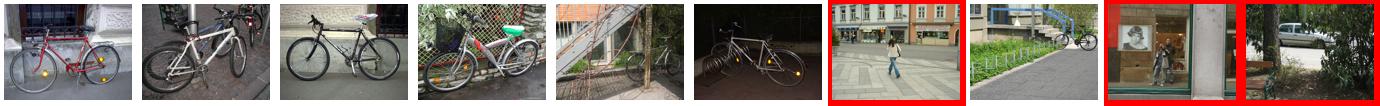}}
\end{tabular}
\label{tab:graz1}
\end{table*}

\begin{table*}[tb]
\caption{Examples of top $10$ retrieval results for \textbf{sailing} (Sports data sets). Red bounding box indicates retrieval error}
\begin{tabular}{ccc}
  & \algoname & \adjustbox{valign=m}{\includegraphics[width=0.8\textwidth,clip]{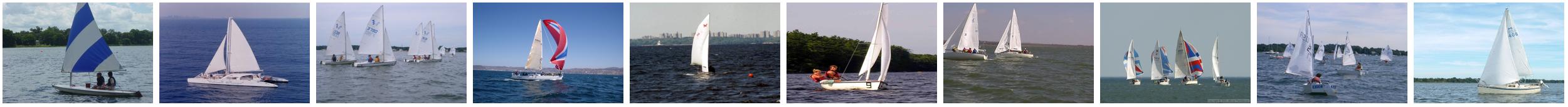}} \\
  & AGH \cite{Liu2011Hashing} & \adjustbox{valign=m}{\includegraphics[width=0.8\textwidth,clip]{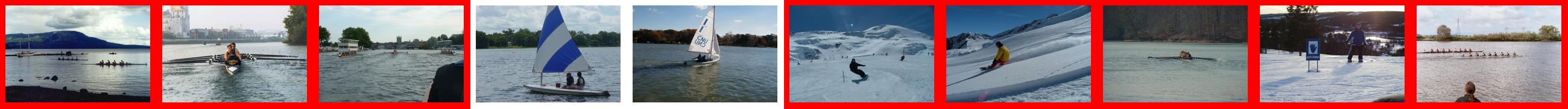}} \\
  \multirow{7}{*}{\includegraphics[width=0.1\textwidth,clip]{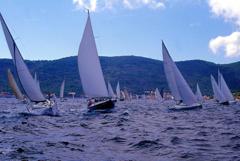}}   & LCH \cite{Zhang2010Laplacian} & \adjustbox{valign=m}{\includegraphics[width=0.8\textwidth,clip]{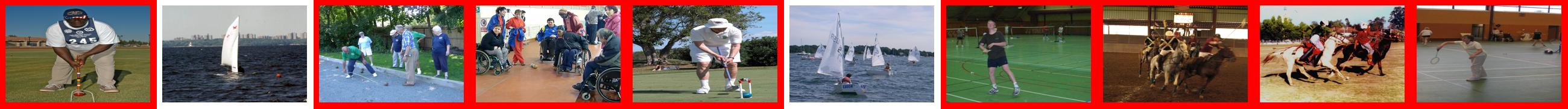}} \\
  & LSI \cite{Deerwester1990Indexing} & \adjustbox{valign=m}{\includegraphics[width=0.8\textwidth,clip]{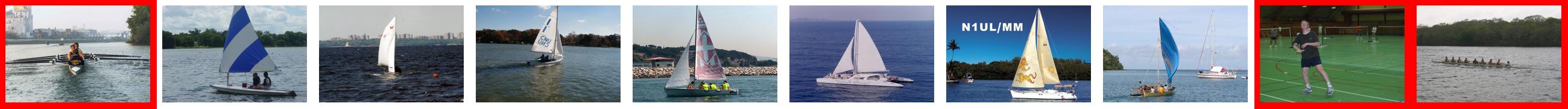}} \\
  & SH \cite{Weiss2008Spectral} & \adjustbox{valign=m}{\includegraphics[width=0.8\textwidth,clip]{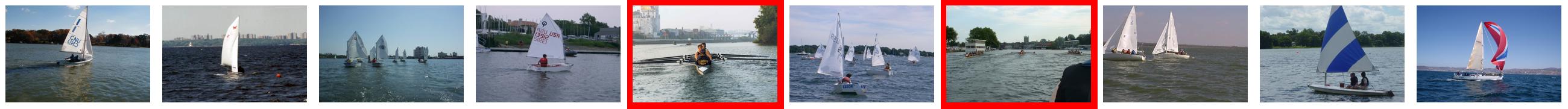}} \\
  & SphH \cite{Heo2012Spherical} & \adjustbox{valign=m}{\includegraphics[width=0.8\textwidth,clip]{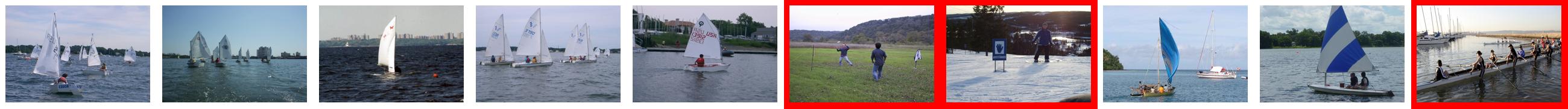}} \\
   & STH \cite{Zhang2010Self}  & \adjustbox{valign=m}{\includegraphics[width=0.8\textwidth,clip]{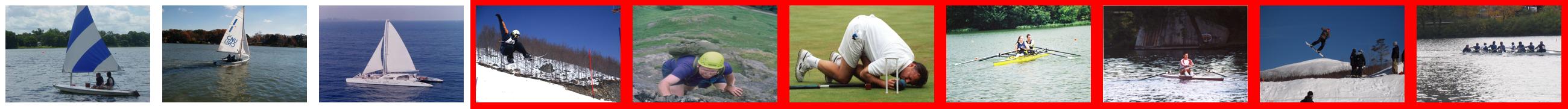}}
\end{tabular}
\label{tab:sports1}
\end{table*}

\section{Conclusion}
We have proposed a novel learning-based approach to building compact binary embeddings.
The method is distinguished from other embedding techniques in that it exploits
extra information in the form of a set of class labels and
pair-wise proximity on the training data set.
Although this learning-based approach is relatively computationally expensive,
it needs only be carried out once offline.
Exploiting  previously ignored class-label information allows the method
to generate embedded features which outperform their original high-dimensional counterparts.
Experimental results demonstrate the effectiveness of our approach for both
parametric and non-parametric models.
Future works include applying the proposed formulation to multi-label classification,
building hierarchical binary output codes and automatically generating the class
labels for the training set.

\appendices

\section{General convex loss with arbitrary regularization}
In this appendix, we generalize our approach to arbitrary
regularization.
We first consider the logistic loss with $\ell_\infty$ penalty.
The learning problem for the logistic loss in the  $\ell_\infty$ regularization
framework can be expressed as
\begin{align}
    \label{EQ:gen1}
        \min_{ \bW, \brho }   \;
        &
        {\textstyle \sum}_{\iprime,r} \log \left( 1 + \exp(-\rho_{\iprime,r}) \right)
          + \nu  \| \bW \|_\infty   \quad   \\ \notag
        \st \; &
        \rho_{\iprime,r} = \bair \bw_{\yiprime} - \bair \bw_{r},
        \forall \iprime, r; \; \bW \geq 0.
\end{align}
Its Lagrange dual can be written as
\begin{align}
    \label{EQ:gen2}
        \max_{ \bU, \bQ }   \quad
        &
        - {\textstyle \sum}_{\iprime,r}
            \Bigl[ \uir \log \left( \uir \right)
            + \\ \notag
            &
            \left(1 - \uir \right)
            \log\left( 1 - \uir \right) \Bigr] + \lambda \| \bQ \|_1
         \\ \notag
        \st \quad &
        {\textstyle  \sum}_{\iprime}
            \bigl[ \delta_{r,\yiprime} \left( \textstyle \sum_{l} \uil \right)
              - \uir \bigr]
                    \bair \leq \nu \bq_{:r}, \forall r;
\end{align}
$\bq_{:r}$ denotes the $r$-th column of $\bQ$ and
$\bQ \geq 0$.
Here the sub-problem for generating the most violated constraint is
\begin{align}
\label{EQ:gen3}
    h^{\ast}(\cdot) = \argmax_{h(\cdot)\in \mathcal{H}, r}
    \;
    {\textstyle \sum}_{\iprime} ( \delta_{r,\yiprime}
     \left( \textstyle \sum_{l} \uil \right)
     -\uir ) \air.
\end{align}
Although the subproblem for generating the binary function
(weak learner) for both $\ell_1$ and $\ell_\infty$ penalties
are the same, the solution to the primal variables, $\bW$,
are different.
In our implementation, we solve the primal problem because the size
of the primal problem is much smaller than the size of the dual problem.

Let us assume any general convex loss functions and arbitrary
regularization penalties, the optimization problem can be rewritten as,
    \begin{align}
        \label{EQ:gen4}
            \min_{ \bW, \brho } \;
            &
            {\textstyle \sum}_{\iprime, r}  \Theta ( \rho_{\iprime,r} ) +
                \nu \Omega( \bW )
            \\ \notag
            \; \st \; &
            \rho_{\iprime,r} = \bair \bw_{\yiprime} - \bair \bw_{r},
            \forall \iprime, r;       \ \text{and} \
            \bW \geq 0.
    \end{align}
    The Lagrangian of \eqref{EQ:gen4} is
    \begin{align}
        \Lag &= {\textstyle \sum}_{\iprime, r}  \Theta  \left( \rho_{\iprime,r} \right) +
                \nu \Omega( \bW )    \\ \notag
            & \; - {\textstyle \sum}_{\iprime,r} \uir
            ( \rho_{\iprime,r} - \bair \bw_{\yiprime}
      + \bair \bw_{r})
             - \trace ( \bP^\T \bW ),
    \end{align}
    Following our derivation for the $\ell_1$ regularized logistic loss,
    the Lagrange dual can be written as,
    \begin{align}
            \min_{ \bU,\bQ }   \;
            &
            {\textstyle \sum}_{\iprime, r}
               \Theta ^ {\ast}(- \uir ) + \lambda \Omega^{\ast}( \bQ )
             \\ \notag
            \st \; &
            {\textstyle  \sum}_{\iprime}
            \bigl[ \delta_{r,\yiprime} \left( \textstyle \sum_{l} \uil \right)
              - \uir \bigr]
                    \bair \leq \nu \bq_{:r}, \forall r;
    \end{align}
    $\bQ \geq 0$.
    Here $ \Theta^{\ast}(\cdot)$ is the Fenchel dual function of $\Theta(\cdot)$
    and $\Omega^{\ast}(\cdot)$ is the Fenchel conjugate of $\Omega(\cdot)$.

\bibliographystyle{ieee}
\bibliography{draft}

\begin{thebibliography}{10}

\bibitem{Andoni2008Near}
A.~Andoni and P.~Indyk,
\newblock ``Near-optimal hashing algorithms for approximate nearest neighbor in
  high dimensions,''
\newblock {\em Communications of the ACM}, vol. 51, no. 1, pp. 117–122, 2008.

\bibitem{Liu2012Spline}
Y.~Liu, F.~Wu, Y.~Yang, Y.~Zhuang, and A.~G. Hauptmann,
\newblock ``Spline regression hashing for fast image search,''
\newblock {\em {IEEE} Trans. Image Proc.}, vol. 21, no. 10, pp. 4480--4491,
  2012.

\bibitem{Weiss2008Spectral}
Y.~Weiss, A.~Torralba, and R.~Fergus,
\newblock ``Spectral hashing,''
\newblock in {\em Proc. Adv. Neural Inf. Process. Syst.}, 2008.

\bibitem{Demiriz2002LPBoost}
A.~Demiriz, K.~P. Bennett, and J.~Shawe-Taylor,
\newblock ``Linear programming boosting via column generation,''
\newblock {\em Mach. Learn.}, vol. 46, no. 1-3, pp. 225--254, 2002.

\bibitem{Shen2012Positive}
C.~Shen, J.~Kim, L.~Wang, and A.~van~den Hengel,
\newblock ``Positive semidefinite metric learning using boosting-like
  algorithms,''
\newblock {\em J. Mach. Learn. Res.}, vol. 13, pp. 1007--1036, 2012.

\bibitem{Datar2004Locality}
M.~Datar, N.~Immorlica, P.~Indyk, and V.~Mirrokni,
\newblock ``Locality-sensitive hashing scheme based on p-stable
  distributions,''
\newblock in {\em Proc. of Symp. on Comp. Geometry}, 2004.

\bibitem{Ke2004Efficient}
Y.~Ke, R.~Sukthankar, and L.~Huston,
\newblock ``Efficient near-duplicate detection and sub-image retrieval,''
\newblock in {\em Proc. of ACM Multimedia}, 2004.

\bibitem{Chum2007Scalable}
O.~Chum, J.~Philbin, M.~Isard, and A.~Zisserman,
\newblock ``Scalable near identical image and shot detection,''
\newblock in {\em Proc. of Int. Conf. on Image and Video Retrieval}, 2007.

\bibitem{Frome2005Object}
A.~Frome and J.~Malik,
\newblock {\em Object Recognition Using Locality-Sensitive Hashing of Shape
  Contexts},
\newblock The MIT Press, 2005.

\bibitem{Liu2011Hashing}
W.~Liu, J.~Wang, S.~Kumar, and S.-F. Chang,
\newblock ``Hashing with graphs,''
\newblock in {\em Proc. Int. Conf. Mach. Learn.}, 2011.

\bibitem{Heo2012Spherical}
J.~Heo, Y.~Lee, J.~He, S.~F. Chang, and S.~E. Yoon,
\newblock ``Spherical hashing,''
\newblock in {\em Proc. IEEE Conf. Comp. Vis. Patt. Recogn.}, 2012.

\bibitem{Lazebnik2006Beyond}
S.~Lazebnik, C.~Schmid, and J.~Ponce,
\newblock ``Beyond bags of features: Spatial pyramid matching for recognizing
  natural scene categories,''
\newblock in {\em Proc. IEEE Conf. Comp. Vis. Patt. Recogn.}, 2006.

\bibitem{Torresani2010Efficient}
L.~Torresani, M.~Szummer, and A.~Fitzgibbon,
\newblock ``Efficient object category recognition using classemes,''
\newblock in {\em Proc. Eur. Conf. Comp. Vis.}, 2010.

\bibitem{Li2010Bank}
L.~J. Li, H.~Su, L.~Fei-Fei, and E.~P. Xing,
\newblock ``Object bank: A high-level image representation for scene
  classification \& semantic feature sparsification,''
\newblock in {\em Proc. Adv. Neural Inf. Process. Syst.}, 2010.

\bibitem{Bergamo2011PICODES}
A.~Bergamo, L.~Torresani, and A.~Fitzgibbon,
\newblock ``Picodes: Learning a compact code for novel-category recognition,''
\newblock in {\em Proc. Adv. Neural Inf. Process. Syst.}, 2011.

\bibitem{Bergamo2012MetaClass}
A.~Bergamo and L.~Torresani,
\newblock ``Meta-class features for large-scale object categorization on a
  budget,''
\newblock in {\em Proc. IEEE Conf. Comp. Vis. Patt. Recogn.}, 2012.

\bibitem{Bengio2010Label}
S.~Bengio, J.~Weston, and D.~Grangier,
\newblock ``Label embedding trees for large multi-class tasks,''
\newblock in {\em Proc. Adv. Neural Inf. Process. Syst.}, 2010.

\bibitem{Boiman2008Defense}
O.~Boiman, E.~Shechtman, and M.~Irani,
\newblock ``In defense of nearest-neighbor based image classification,''
\newblock in {\em Proc. IEEE Conf. Comp. Vis. Patt. Recogn.}, 2008.

\bibitem{Boureau2010Learning}
Y.~L. Boureau, F.~Bach, Y.~LeCun, and J.~Ponce,
\newblock ``Learning mid-level features for recognition,''
\newblock in {\em Proc. IEEE Conf. Comp. Vis. Patt. Recogn.}, 2010.

\bibitem{Yang2009Linear}
J.~Yang, K.~Yu, Y.~Gong, and T.~S. Huang.,
\newblock ``Linear spatial pyramid matching using sparse coding for image
  classification,''
\newblock in {\em Proc. IEEE Conf. Comp. Vis. Patt. Recogn.}, 2009.

\bibitem{Wang2010Locality}
J.~Wang, J.~Yang, K.~Yu, F.~Lv, T.~Huang, and Y.~Gong,
\newblock ``Locality-constrained linear coding for image classification,''
\newblock in {\em Proc. IEEE Conf. Comp. Vis. Patt. Recogn.}, 2010.

\bibitem{Wang2013Linear}
Z.~Wang, J.~Feng, S.~Yan, and H.~Xi,
\newblock ``Linear distance coding for image classification,''
\newblock {\em {IEEE} Trans. Image Proc.}, vol. 22, no. 2, pp. 537--548, 2013.

\bibitem{Behmo2010Towards}
R.~Behmo, P.~Marcombes, A.~Dalalyan, and V.~Prinet,
\newblock ``Towards optimal naive bayes nearest neighbor,''
\newblock in {\em Proc. Eur. Conf. Comp. Vis.}, 2010.

\bibitem{Tuytelaars2011NBNN}
T~Tuytelaars, M.~Fritz, K.~Saenko, and T.~Darrell,
\newblock ``The {NBNN} kernel,''
\newblock in {\em Proc. IEEE Int. Conf. Comp. Vis.}, 2011.

\bibitem{McCann2012Local}
S.~McCann and D.~G. Lowe,
\newblock ``Local naive bayes nearest neighbor for image classification,''
\newblock in {\em Proc. IEEE Conf. Comp. Vis. Patt. Recogn.}, 2012.

\bibitem{Wang2010Image}
Z.~Wang, Y.~Hu, and L.-T. Chia,
\newblock ``{Image-to-Class} distance metric learning for image
  classification,''
\newblock in {\em Proc. Eur. Conf. Comp. Vis.}, 2010.

\bibitem{Lowe2004Distinctive}
David~G. Lowe,
\newblock ``Distinctive image features from scale-invariant keypoints,''
\newblock {\em Int. J. Comp. Vis.}, vol. 60, no. 2, pp. 91--110, 2004.

\bibitem{Zhang2013Edge}
S.~Zhang, Q.~Tian, K.~Lu, Q.~Huang, and W.~Gao,
\newblock ``Edge-{SIFT}: Discriminative binary descriptor for scalable
  partial-duplicate mobile search,''
\newblock {\em {IEEE} Trans. Image Proc.}, vol. 22, no. 7, pp. 2889--2902,
  2013.

\bibitem{Bay2006SURF}
H.~Bay, T.~Tuytelaars, and L.~Van Gool,
\newblock ``{SURF}: Speeded up robust features,''
\newblock in {\em Proc. Eur. Conf. Comp. Vis.}, 2006.

\bibitem{Kira1992Practical}
K.~Kira and L.~A. Rendell,
\newblock ``A practical approach to feature selection,''
\newblock in {\em Proc. Int. Conf. Mach. Learn.}, 1992.

\bibitem{Sun2010Local}
Y.~Sun, S.~Todorovic, and S.~Goodison,
\newblock ``Local-learning-based feature selection for high-dimensional data
  analysis,''
\newblock {\em {IEEE} Trans. Pattern Anal. Mach. Intell.}, vol. 32, no. 9, pp.
  1610--1626, 2010.

\bibitem{Weinberger2006Distance}
K.~Q. Weinberger, J.~Blitzer, and L.~K. Saul,
\newblock ``Distance metric learning for large margin nearest neighbor
  classification,''
\newblock in {\em Proc. Adv. Neural Inf. Process. Syst.}, 2006.

\bibitem{Maaten2012Stochastic}
L.~J.~P. van~der Maaten and K.~Q. Weinberger,
\newblock ``Stochastic triplet embedding,''
\newblock in {\em IEEE Intl. Workshop on Mach. Learn. for Sig. Proc.}, 2012.

\bibitem{Liu2007Gradient}
X.~Liu and T.~Yu,
\newblock ``Gradient feature selection for online boosting,''
\newblock in {\em Proc. IEEE Int. Conf. Comp. Vis.}, 2007.

\bibitem{Zhu1997Algorithm}
C.~Zhu, R.~H. Byrd, P.~Lu, and J.~Nocedal,
\newblock ``Algorithm 778: L-bfgs-b: Fortran subroutines for large-scale
  bound-constrained optimization,''
\newblock {\em ACM Trans. Math. Software}, vol. 23, no. 4, pp. 550--560, 1997.

\bibitem{Le2011Optimization}
Q.V. Le, J.~Ngiam, A.~Coates, A.~Lahiri, B.~Prochnow, and A.Y. Ng.,
\newblock ``On optimization methods for deep learning.,''
\newblock in {\em Proc. Int. Conf. Mach. Learn.}, 2011.

\bibitem{Norouzi2012Hamming}
M.~Norouzi, D.~Fleet, and R.~Salakhutdinov,
\newblock ``Hamming distance metric learning,''
\newblock in {\em Proc. Adv. Neural Inf. Process. Syst.}, 2012.

\bibitem{Vedaldi2010Efficient}
A.~Vedaldi and A.~Zisserman,
\newblock ``Efficient additive kernels via explicit feature maps,''
\newblock in {\em Proc. IEEE Conf. Comp. Vis. Patt. Recogn.}, 2010.

\bibitem{Vedaldi08Vlfeat}
A.~Vedaldi and B.~Fulkerson,
\newblock ``{VLFeat}: An open and portable library of computer vision
  algorithms,'' \url{http://www.vlfeat.org/}, 2008.

\bibitem{Aly2011Indexing}
M.~Aly, M.~Munich, and P.~Perona,
\newblock ``Indexing in large scale image collections: Scaling properties and
  benchmark,''
\newblock in {\em IEEE Workshop on Apps. of Comp. Vis.}, 2011.

\bibitem{Viola2004Robust}
P.~Viola and M.~J. Jones,
\newblock ``Robust real-time face detection,''
\newblock {\em Int. J. Comp. Vis.}, vol. 57, no. 2, pp. 137--154, 2004.

\bibitem{Munder2006Experimental}
S.~Munder and D.~M. Gavrila,
\newblock ``An experimental study on pedestrian classification,''
\newblock {\em {IEEE} Trans. Pattern Anal. Mach. Intell.}, vol. 28, no. 11, pp.
  1863--1868, 2006.

\bibitem{Dalal2005Histograms}
N.~Dalal and B.~Triggs,
\newblock ``Histograms of oriented gradients for human detection,''
\newblock in {\em Proc. IEEE Conf. Comp. Vis. Patt. Recogn.}, 2005.

\bibitem{Wojek2009Multi}
C.~Wojek, S.~Walk, and B.~Schiele,
\newblock ``Multi-cue onboard pedestrian detection,''
\newblock in {\em Proc. IEEE Conf. Comp. Vis. Patt. Recogn.}, 2009.

\bibitem{Dollar2012Pedestrian}
P.~Doll\'ar, C.~Wojek, B.~Schiele, and P.~Perona,
\newblock ``Pedestrian detection: An evaluation of the state of the art,''
\newblock {\em {IEEE} Trans. Pattern Anal. Mach. Intell.}, vol. 34, no. 4, pp.
  743--761, 2012.

\bibitem{Torralba2003Context}
A.~Torralba, K.~P. Murphy, W.~T. Freeman, and M.~A. Rubin,
\newblock ``Context-based vision system for place and object recognition,''
\newblock in {\em Proc. IEEE Int. Conf. Comp. Vis.}, 2003.

\bibitem{Bo2011Object}
L.~Bo, K.~Lai, X.~Ren, and D.~Fox,
\newblock ``Object recognition with hierarchical kernel descriptors,''
\newblock in {\em Proc. IEEE Conf. Comp. Vis. Patt. Recogn.}, 2011.

\bibitem{Zhang2010Self}
D.~Zhang, J.~Wang, D.~Cai, and J.~Lu,
\newblock ``Self-taught hashing for fast similarity search,''
\newblock in {\em ACM SIGIR}, 2010.

\bibitem{Zhang2010Laplacian}
D.~Zhang, J.~Wang, D.~Cai, and J.~Lu.,
\newblock ``Laplacian co-hashing of terms and documents,''
\newblock in {\em European Conf. on IR Res.}, 2010.

\bibitem{Wu2011CENTRIST}
J.~Wu and J.~M. Rehg,
\newblock ``{CENTRIST}: A visual descriptor for scene categorization,''
\newblock {\em {IEEE} Trans. Pattern Anal. Mach. Intell.}, vol. 33, no. 8, pp.
  1489--1501, 2011.

\bibitem{Fan2008Liblinear}
R.-E. Fan, K.-W. Chang, C.-J. Hsieh, X.-R. Wang, and C.-J. Lin,
\newblock ``{LIBLINEAR}: A library for large linear classification,''
\newblock {\em J. Mach. Learn. Res.}, vol. 9, pp. 1871--1874, 2008.

\bibitem{Weiss2012Multi}
Y.~Weiss, R.~Fergus, and A.~Torralba,
\newblock ``Multidimensional spectral hashing,''
\newblock in {\em Proc. Eur. Conf. Comp. Vis.}, 2012.

\bibitem{Deerwester1990Indexing}
S.~C. Deerwester, S.~T. Dumais, T.~K. Landauer, G.~W. Furnas, and R.~A.
  Harshman,
\newblock ``Indexing by latent semantic analysis,''
\newblock {\em Journal American Socity for Information Science}, vol. 41, no.
  6, pp. 391--407, 1990.

\end{thebibliography}

\end{document}